\documentclass{article}

\usepackage[preprint]{neurips_2026}
\usepackage[utf8]{inputenc} 
\usepackage[T1]{fontenc}    
\usepackage{hyperref}       
\usepackage{url}            
\usepackage{booktabs}       
\usepackage{amsfonts}       
\usepackage{nicefrac}       
\usepackage{microtype}      
\usepackage{xcolor}         
\usepackage{graphicx}
\usepackage{amsmath}
\usepackage{amsfonts}
\usepackage{subcaption}
\usepackage{caption}
\usepackage{float}
\usepackage{mathtools}
\usepackage{extpfeil} 
\usepackage{natbib}
\usepackage{wrapfig}
\usepackage{overpic}
\usepackage{xspace}
\usepackage{enumitem}
\usepackage{xurl}

\usepackage{multirow}
\usepackage{multicol}
\usepackage{siunitx}
\usepackage{bm}
\usepackage{listings}
\usepackage{tikz}
\usepackage{tabularx}   
\usepackage{tabularray} 

\usepackage{authblk}
\usepackage{bbm}

\usepackage{xcolor}
\usepackage{listings}

\definecolor{pddlKeyword}{RGB}{0, 76, 153}
\definecolor{popddlKeyword}{RGB}{117, 63, 152}
\definecolor{probKeyword}{RGB}{190, 100, 30}
\definecolor{pddlComment}{RGB}{80, 120, 80}
\definecolor{pddlString}{RGB}{150, 60, 60}
\definecolor{pddlBg}{RGB}{248, 248, 248}

\lstdefinelanguage{POPDDL}{
  morekeywords={
    define,domain,problem,
    requirements,types,constants,predicates,functions,
    action,parameters,precondition,effect,
    objects,init,goal,metric,
    and,not,when
  },
  morekeywords=[2]{
    observables,observation,condition,distribution,init-belief,joint
  },
  morekeywords=[3]{
    probabilistic,increase,decrease,maximize,total-reward
  },
  sensitive=false,
  morecomment=[l]{;;},
  morestring=[b]"
}

\lstdefinestyle{popddlstyle}{
  language=POPDDL,
  backgroundcolor=\color{pddlBg},
  basicstyle=\ttfamily\scriptsize,
  keywordstyle=\color{pddlKeyword}\bfseries,
  keywordstyle=[2]\color{popddlKeyword}\bfseries,
  keywordstyle=[3]\color{probKeyword}\bfseries,
  commentstyle=\color{pddlComment}\itshape,
  stringstyle=\color{pddlString},
  numbers=none,
  numberstyle=\tiny\color{gray},
  stepnumber=1,
  numbersep=8pt,
  showstringspaces=false,
  breaklines=true,
  breakatwhitespace=false,
  columns=fullflexible,
  keepspaces=true,
  frame=single,
  rulecolor=\color{gray!40},
  xleftmargin=2.2em,
  framexleftmargin=2.0em,
  aboveskip=0.6em,
  belowskip=0.4em,
  captionpos=b
}

\lstdefinestyle{promptstyle}{
    backgroundcolor=\color{gray!10},
    basicstyle=\ttfamily\tiny,
    frame=single,
    rulecolor=\color{gray!40},
    breaklines=true,
    breakatwhitespace=false,
    columns=fullflexible,
    keepspaces=true,
    numbers=none,
    showstringspaces=false,
    keywordstyle=\color{blue},
    commentstyle=\color{gray},
    stringstyle=\color{red},
    xleftmargin=0.8em,
    framexleftmargin=0.6em,
    aboveskip=0.6em,
    belowskip=0.4em,
    captionpos=b
}

\lstdefinestyle{pythonstyle}{
    language=Python,
    backgroundcolor=\color{gray!10},
    basicstyle=\ttfamily\scriptsize,
    keywordstyle=\color{blue!70!black}\bfseries,
    commentstyle=\color{gray}\itshape,
    stringstyle=\color{red!70!black},
    numbers=none,
    numberstyle=\tiny\color{gray},
    stepnumber=1,
    numbersep=8pt,
    showstringspaces=false,
    breaklines=true,
    breakatwhitespace=false,
    columns=fullflexible,
    keepspaces=true,
    frame=single,
    rulecolor=\color{gray!40},
    xleftmargin=2.2em,
    framexleftmargin=2.0em,
    aboveskip=0.6em,
    belowskip=0.4em,
    captionpos=b,
    tabsize=4
}

\usepackage{siunitx}

\title{PO-PDDL: Learning Symbolic POMDPs from Visual Demonstrations for Robot Planning Under Uncertainty}

\author{
\textbf{Wenjing Tang}\textsuperscript{1,2\S},
\textbf{Xuanjin Jin}\textsuperscript{1,2},
\textbf{Yuan Liu}\textsuperscript{1,2},
\textbf{Renming Huang}\textsuperscript{1,2},
\textbf{Cewu Lu}\textsuperscript{1,2},
\textbf{Panpan Cai}\textsuperscript{1,2\dag} \\
\textsuperscript{1}Shanghai Jiao Tong University \quad
\textsuperscript{2}Shanghai Innovation Institute \\
}

\usepackage{xspace}
\usepackage{hyperref}

\newcommand{\formuname}{PO-PDDL\xspace}

\newcommand{\secref}[1]{Section~\ref{#1}}

\renewcommand{\eqref}[1]{(\ref{#1})}

\newcommand{\figref}[1]{Fig.~\ref{#1}}

\usepackage{etoolbox}
\newtoggle{final}

\usepackage{fontawesome}

\newcommand{\unidomain}{\textit{UniDomain$^*$}\xspace}
\newcommand{\pomdpcoder}{\textit{pomdp\_coder$^*$}\xspace}
\newcommand{\vlmplanner}{\textit{VLM-Planner}\xspace}

\newcommand{\jointobslearn}{Open-Pass\xspace}
\newcommand{\noactiveobs}{No-ActiveObs\xspace}

\newcommand{\taskpushbox}{\textit{push\_box}\xspace}
\newcommand{\taskblock}{\textit{block\_in\_drawer}\xspace}
\newcommand{\taskcup}{\textit{cup\_with\_water}\xspace}
\newcommand{\taskblockcup}{\textit{cups\_then\_blocks}\xspace}
\newcommand{\taskblockcupbox}{\textit{box\_cups\_blocks}\xspace}

\begin{document}

\maketitle

{\renewcommand{\thefootnote}{\S}
 \footnotetext{Wenjing Tang is the visiting students at Shanghai Innovation Institute.}
}
{\renewcommand{\thefootnote}{\dag}%
 \footnotetext{Corresponding author: \texttt{cai\_panpan@sjtu.edu.cn}}
}

\begin{abstract}

Real-world robot task planning must operate under both stochastic action execution and partial observability, yet constructing Partially Observable Markov Decision Process (POMDP) models for real robotics domains remains difficult and labor-intensive. 
We introduce PO-PDDL, a symbolic formulation of POMDPs that preserves the relational structure and LLM-friendly syntax of the Planning Domain Definition Language (PDDL), while explicitly modeling partial observability, stochasticity, and beliefs. 
Building on this formulation, we propose a demonstration-driven pipeline for learning PO-PDDL models. The proposed method reconstructs latent symbolic state trajectories from real-robot execution videos, identifies partial observability via inconsistencies between inferred states and visual observations, and learns stochastic transition and observation models accordingly. 
The resulting PO-PDDL domains are reusable across tasks and enable online belief-space planning under both perception and execution uncertainty. Experiments on real-world long-horizon manipulation tasks show that our method consistently outperforms existing PDDL and POMDP model-learning approaches, achieving robust task planning under uncertainty with significantly lower planning cost.\footnote{Project page: \url{https://roboticsjtu.github.io/PO-PDDL/}.}


\end{abstract}

\section{Introduction}
\label{intro}

Real-world robot task planning must operate under substantial uncertainty arising from both perception and action execution. In practical settings, robots face perceptual uncertainty caused by occlusions, limited sensing, and perception errors, as well as stochastic action outcomes from low-level execution modules such as motion planners and vision-language-action (VLA)~\cite{black2024pi_0, intelligence2025pi_} policies. These uncertainties naturally call for modeling robot decision-making as a Partially Observable Markov Decision Process (POMDP). However, constructing POMDP models for real-world tasks remains challenging: specifying transition, observation, and reward functions requires significant expertise, and learning such models directly from experience—e.g., from visual demonstrations of policy executions—is highly non-trivial.

Existing approaches fall short of addressing this challenge. On the one hand, recent work that learns symbolic planning domains in the Planning Domain Definition Language (PDDL)~\cite{aeronautiques1998pddl, younes2004ppddl1}, such as UniDomain~\cite{yeunidomain}, supports structured reasoning but assumes fully observable and deterministic environments. On the other hand, POMDP formulations can be specified in languages such as POMDPX~\cite{du2010pomdp} or implemented in Python, and recent attempts have explored learning such models from demonstrations using Large Language Models (LLMs)~\cite{kim2026large}. In practice, however, these approaches remain difficult to scale to visual robot domains. They are typically evaluated in simplified low-dimensional settings and struggle with the state abstraction, stochasticity, and partial observability present in real robot demonstrations.

In this work, we propose a new approach to bridge this gap. We first introduce \textbf{\formuname}, an LLM-friendly symbolic POMDP formulation that extends PDDL-style representations with explicit observation modeling and belief specifications. \formuname preserves the structured, relational syntax that enables LLM-based generation, while supporting stochastic transitions and partial observability. Building on this formulation, we develop a learning pipeline that constructs a \formuname domain directly from visual demonstrations of robot policy executions. Our method infers predicate schemas, stochastic operator models, and observation functions by reconstructing hidden symbolic trajectories and identifying inconsistencies between inferred states and visual observations. The learned domain is reusable across tasks in the same environment and supports online planning under uncertainty.

We evaluate our method on real-world robot tasks involving long horizons, inter-action constraints, stochastic skill execution, and partial observability. Results show that our method effectively tackles these uncertainties, achieving a 100\% success rate and the highest cumulative reward across all task levels, and substantially outperforming both other model learning methods and VLM planning.

In summary, we make the following contributions:
(i) We propose \formuname, a symbolic formulation that unifies PDDL-style representations with POMDP modeling of uncertainty;
(ii) We develop a learning pipeline that constructs \formuname domains from visual demonstrations, including stochastic transitions and observation models;
(iii) We introduce a mismatch-based mechanism for identifying partial observability from data;
(iv) We demonstrate that the learned domains enable belief-space planning for long-horizon robot tasks under both perception and execution uncertainty.

\section{Related Work}
\label{related_work}
\subsection{Domain Definition Languages for Uncertainty}
POMDP planning requires explicit transition, observation, reward, and initial-belief models. POMDPX~\cite{du2010pomdp} provides an XML-based representation, but is poorly aligned with LLM generation and often requires manual model engineering. RDDL~\cite{sanner2010relational} offers a Dynamic Bayesian Network (DBN)-based relational formalism, but its compact dependency syntax is difficult for current LLMs to synthesize reliably. These limitations motivate \formuname, which retains PDDL-style relational structure while adding observation and belief specifications for stochastic partially observable planning.
\subsection{LLM-based POMDP Planning}
LLM-based POMDP planning can be broadly categorized into implicit and explicit approaches. Implicit methods use LLMs or VLMs directly for planning through prompting, uncertainty estimation, or reinforcement learning~\cite{sun2024interactive,ren2023robots,mullen2024towards,jiang2024llms,wang2025vagen}. However, their beliefs and dynamics remain implicit, limiting solver-level guarantees and requiring repeated model inference online.
Explicit methods combine LLMs with symbolic or probabilistic planners. LLM-MCTS~\cite{zhao2023large} uses LLM-generated beliefs for Monte Carlo tree search, but handles limited hidden factors and incurs high online LLM cost. Tru-POMDP~\cite{tang2025tru} uses hypothesis trees but relies on manually specified models. VLM-guided POMDPs for object search~\cite{wang2025robust} are specialized to navigation. In contrast, our method learns reusable transition, observation, reward, and initial-belief components from visual demonstrations, enabling solver-based online planning without task-specific manual engineering.
\subsection{Learning Planning Models from Demonstrations}
Learning planning models from demonstrations reduces manual modeling effort. Prior work learns PDDL domains or problem instances from demonstrations~\cite{yeunidomain, la2025end, wang2025unifying} and repairs symbolic models using environmental feedback~\cite{byrnes2025climb,huang2025one,mahdavi2024leveraging,yu2025generating}, but mainly assumes deterministic dynamics and does not learn probabilistic transition or observation models.
Recent methods address partial observability via iterative deterministic replanning~\cite{gong2025zero} or symbolic belief-state planning with coarse three-valued uncertainty~\cite{zhao2025seeing}, but still do not learn full probabilistic transition and observation models for POMDP belief-space planning. Beyond deterministic symbolic learning, PlanU~\cite{deng2025planu} learns stochastic action outcomes, and prior work \cite{curtis2025llm} estimates POMDP transition and observation models, but both are typically evaluated in structured domains with simplified state abstractions. Our method addresses this gap by learning probabilistic partially observable models directly from robot execution videos and using them for closed-loop belief-space planning under transition and observation uncertainty.


\section{Problem Formulation: Partially Observable PDDL}
\label{sec:form}
\subsection{Existing Formulations}

\paragraph{PDDL.}
The Planning Domain Definition Language (PDDL) represents planning problems in a symbolic and factored form. A classical PDDL problem is defined by
$
\langle \mathcal{P}, \mathcal{A}, s_0, \mathcal{G} \rangle,
$
where $\mathcal{P}$ denotes the predicates, $\mathcal{A}$ operators, $s_0$ the initial state, and $\mathcal{G}$ the goal condition.
Each operator $a \in \mathcal{A}$ is defined by:
$
a = \langle \mathrm{pre}(a), \mathrm{eff}(a), R(a)\rangle,
$
where $\mathrm{pre}(a)$ specifies applicability conditions, $\mathrm{eff}(a)$ defines deterministic or stochastic state transitions and $R(a)$ represents the reward (in PPDDL~\cite{younes2004ppddl1}). In practice, $\mathcal{P}$ and $\mathcal{A}$ are specified in a PDDL \textit{domain file}, while $s_0$ and $\mathcal{G}$ are defined in a PDDL \textit{problem file}. The former is shared across tasks within the same domain.



\paragraph{POMDP.}
A Partially Observable Markov Decision Process (POMDP) models decision-making under uncertainty as
$
\langle S, A, O, T, Z, R, b_0 \rangle,
$
where $S$ is the state space, $A$ the action space, $O$ the observation space, $T$ the transition function, $Z$ the observation function, $R$ the reward, and $b_0$ the initial belief. Both $T$ and $Z$ are specified as conditional probability distributions.
POMDPs handle stochastic dynamics and partial observability through belief updates and belief-space planning. 

PDDL and POMDPs therefore offer complementary strengths: PDDL is symbolic, interpretable, and LLM-friendly, while POMDPs provide principled uncertainty modeling. This motivates a unified formulation that preserves the relational structure of PDDL while incorporating stochastic transitions, observations, and belief-state planning.




\subsection{PO-PDDL}

We propose \textbf{Partially Observable PDDL (\formuname)}, a symbolic formulation of POMDPs that preserves the structured syntax of PDDL while explicitly modeling partial observability. A \formuname model is defined as $\mathcal{M}=\langle \mathcal{P},\mathcal{O},\mathcal{A},\mathcal{Z}, b_0,\mathcal{G}\rangle$, where $\mathcal{P}$ is a set of predicate schemas, $\mathcal{O}$ is a set of observable schemas, $\mathcal{A}$ is a set of stochastic operators, $\mathcal{Z}$ is a symbolic observation model, $b_0$ is an initial belief, and $\mathcal{G}$ is a goal condition.


A predicate schema $p\in\mathcal{P}$ induces grounded predicates $g=p(\bar{x})$ by assigning objects $\bar{x}$ to its typed arguments; let $\mathcal{G}_{\mathcal{P}}$ be the set of all such predicates. A fully observed symbolic state is a complete assignment $s:\mathcal{G}_{\mathcal{P}}\rightarrow\{0,1\}$, with $g\in s$ denoting $s(g)=1$, while a partially observed state is a partial assignment $\hat{s}:\mathcal{G}_{\mathcal{P}}\rightarrow\{0,1,?\}$, where ``$?$'' denotes an unknown value. An observable schema $o_p\in\mathcal{O}$ is associated with a predicate schema $p$, representing perception outcomes. 
Once grounded, $\mathcal{P}$ and $\mathcal{O}$ define the POMDP state space $S$ and observation space $O$, respectively.

Each operator $a\in\mathcal{A}$ has a precondition $\mathrm{pre}(a)$, stochastic effects $\mathrm{eff}(a)$ over grounded predicates, and a reward $R(a)$. Its stochastic effects induce a transition distribution $T(s'|s,a)$ over complete predicate assignments. The observation model $\mathcal{Z}$ is specified as conditional symbolic rules: each rule maps a logical condition over grounded predicates to a probability distribution over observable values. Once grounded, $\mathcal{A}$ defines the POMDP action space $A$, transition function $T$ and reward function $R$, while $\mathcal{Z}$ represents the POMDP observation function $Z$.

The initial belief $b_0$ is a probability distribution over complete symbolic states $s$, rather than over direct visual observations. 
The goal $\mathcal{G}$ is a logical condition over grounded predicates, or equivalently a set of satisfying complete assignments, and induces a terminal reward.

Similarly to PDDL, $\mathcal{P}$, $\mathcal{O}$, $\mathcal{A}$, and $\mathcal{Z}$ are specified in a \textit{domain file}, while $b_0$ and $\mathcal{G}$ are defined in a \textit{problem file}. 
See Appendix~\ref{app:tiger} for an example of \formuname files.

\section{Learning a \formuname Domain from Demonstrations}
\label{sec:learn}

\begin{figure}
    \centering
    \includegraphics[width=1.0\linewidth]{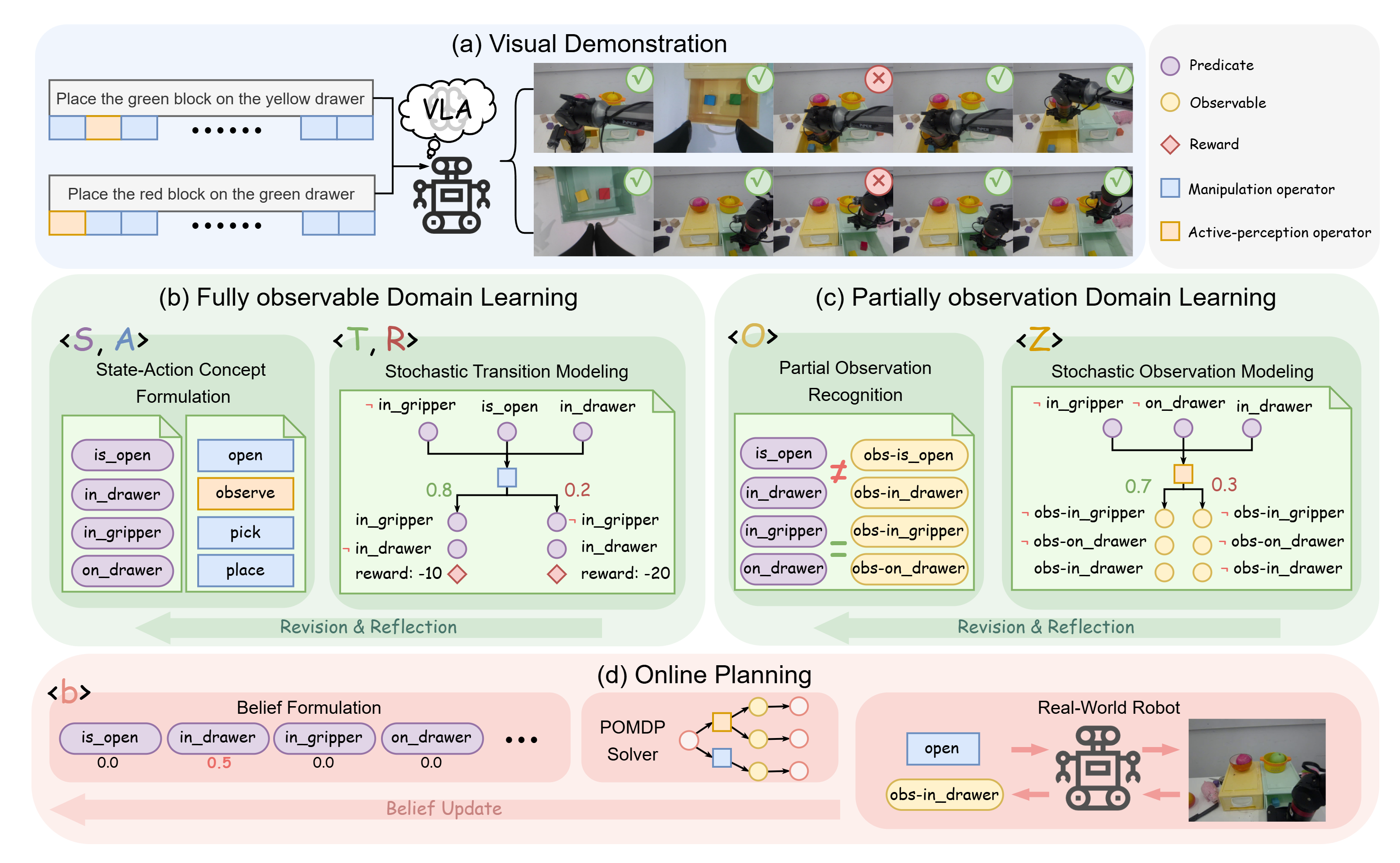}
    \vspace*{-0.5em}
    \caption{    
     Overview of our method. See detailed descriptions in \secref{sec:learn} and \secref{sec:plan}.}
    \label{fig:pipeline}
    \vspace*{-1.5em}
\end{figure}

A \formuname domain is tied to a specific scene and a fixed low-level execution stack, e.g., the low-level execution policy and the perception module, but is shared across tasks.
To learn such a domain, we assume a visual-language demonstration dataset $\mathcal{D}=\{D_i\}_{i=1}^K$, where each episode $D_i$ rolls out a multi-step high-level task (\figref{fig:pipeline}a). 
The high-level task is given as an ordered sequence of natural-language subtasks, each executed by the low-level policy. During the rollout, if a subtask fails, a recovery subtask is inserted, until the overall task is completed.


The resulting episode of task $i$ is represented as $D_i=(\mathcal{V}_i,\mathcal{L}_i)$. This includes visual observations $\mathcal{V}_i=\{(V^{\mathrm{h}}_{i,t}, V^{\mathrm{w}}_{i,t})\}_{t=0}^{T_i}$, comprising synchronized RGB streams from head and wrist cameras. Language annotations $\mathcal{L}_i=(L_i,\{(L_{i,t}, X_{i,t}, Y_{i,t})\}_{t=1}^{T_i})$ contain a task instruction $L_i$ and step-level annotations---$L_{i,t}$ is the subtask instruction at step $t$, $X_{i,t}$ the VLM-generated scene description, and $Y_{i,t}\in\{0,1\}$ the execution outcome. Videos are segmented into subtask-aligned clips.
Our goal is to learn a \formuname domain $\mathcal{M}=\langle \mathcal{P},\mathcal{O},\mathcal{A},\mathcal{Z}\rangle$ from $\mathcal{D}=\{D_i\}_{i=1}^K$, where each problem instance in this domain induces a POMDP model $\langle S,A,O,T,Z,R\rangle$.

The learning pipeline proceeds in two stages. We first learn a fully observable symbolic model $(\mathcal{P},\mathcal{A})$ by reconstructing hidden state trajectories (\figref{fig:pipeline}b). We then identify partially observable components $(\mathcal{O},\mathcal{Z})$ by analyzing contradictions between reconstructed states and visual observations (\figref{fig:pipeline}c), yielding a complete \formuname domain.

\subsection{Fully Observable Domain Learning}
The following pipeline learns predicate schemas and operator interfaces of a fully observable domain.

\subsubsection{State-Action Concept Formulation}
We begin by inducing the symbolic vocabulary from visual-language demonstrations $\mathcal{D}$.



\begin{enumerate}[
    label=(\alph*),
    wide=0pt,
    leftmargin=0pt,
    labelsep=0.4em,
    itemsep=0pt,
    parsep=0pt,
    topsep=0pt,
    partopsep=0pt
]
\item \textit{State concepts}. 
An LLM takes as input the task instruction $L_i$, subtask instructions $\{L_{i,t}\}$, and VLM-generated scene descriptions $\{X_{i,t}\}$. It generates a set of \textit{types} that capture static attributes of entities (e.g., object classes), and a set of \textit{predicates} $\mathcal{P}$ that represent variable attributes, including object properties and spatial relations that evolve across time and are necessary for explaining task progression  (e.g., ``object\_on\_table'', ``container\_is\_open'').
\item \textit{Action concepts}. 
The LLM abstracts subtask instructions $\{L_{i,t}\}$ into a set of operator interfaces $\tilde{\mathcal{A}}$, where each interface specifies an operator name and parameter signature, but without preconditions or effects, analogous to function interfaces without implementation. Operators are further classified as either \textit{manipulation operators}, which intend to change the environment state, or \textit{active-perception operators}, which aim at gathering information about unknown predicate values.
\end{enumerate}

\subsubsection{Stochastic Transition Modeling}

Given predicate schemas $\mathcal{P}$ and operator interfaces $\tilde{\mathcal{A}}$, we now construct stochastic operator models.

\begin{enumerate}[
    label=(\alph*),
    wide=0pt,
    leftmargin=0pt,
    labelsep=0.4em,
    itemsep=2pt,
    parsep=0pt,
    topsep=0pt,
    partopsep=0pt
]

\item \textit{Goal and action grounding.} 
For each episode $D_i$, an LLM formulates a symbolic goal $\mathcal{G}_i$ according to the task instruction $L_i$, using $\mathcal{P}$ as the vocabulary. Subtask instructions $\{L_{i,t}\}_{t=1}^{T_i}$ are then mapped to an action sequence $\mathbf{a}_i=(a_{i,1},\dots,a_{i,T_i})$ by selecting operator interfaces from $\tilde{\mathcal{A}}$ and assigning parameters (e.g., typed objects). Next, we derive state trajectories.

\item \textit{State trajectory reconstruction.} 
Due to partial observability, scene descriptions $\{X_{i,t}\}$ may not reflect ground-truth states. We propose forward-backward inference. 
In the forward pass, for each action $a_{i,t}$, an LLM infers its symbolic effects $\epsilon_{i,t}=(\epsilon^+_{i,t}, \epsilon^-_{i,t})$ from visual observations before and after execution, where $\epsilon^+_{i,t}$ and $\epsilon^-_{i,t}$ denote added and removed grounded predicates.
Next, given $\{\mathbf{a}_i\}$ and  $\{\epsilon_{i,t}\}$, together with scene descriptions $\{X_{i,t}\}_{t=0}^{T_i}$ and the goal $\mathcal{G}_i$, the LLM performs backward inference to derive a ground-truth initial state $s_{i,0}$. 
Then, starting from $s_{i,0}$, sequentially applying effects $\{\epsilon_{i,t}\}$ yields the reconstructed trajectory $\{s_{i,t}\}_{t=0}^{T_i}$. 
Finally, we enforce goal consistency by requiring $s_{i,T_i}\models \mathcal{G}_i$; if violated, the LLM revises $s_{i,0}$, $\{\epsilon_{i,t}\}$, or $\mathcal{G}_i$ until consistency is achieved.

\item \textit{Stochastic operator modeling.} 
Now, we can model preconditions and stochastic effects of operators.
For each operator $\alpha\in\tilde{\mathcal{A}}$, its precondition is obtained as predicate assignments that are consistently seen in all pre-states $\{s_{i,t}\}$ where $\alpha$ is applied. Operator effects are obtained by clustering all seen effects $\{\epsilon_{i,t}\}$ into distinct modes and assigning probabilities to each mode based on empirical frequencies.
The reward of each operator is defined as the negative execution cost, which is proportional to the average duration of its corresponding subtask. 
The above procedure transforms abstract operator interfaces $\tilde{\mathcal{A}}$ into fully-specified operators $\mathcal{A}$.
\end{enumerate}

\subsubsection{Domain Review}


To ensure quality, we apply a VLM-based verification loop to refine the learned operator set $\mathcal{A}$. For each effect mode of an operator $\alpha\in\mathcal{A}$, we sample a representative subtask instance $(i,t)$ and prompt the VLM to check whether its inferred predicates, preconditions, and effects are consistent with the visual observations $(V^{\mathrm{h}}_{i,t}, V^{\mathrm{w}}_{i,t})$. If any mismatch is found, the VLM revises the corresponding symbolic descriptions of that step. The affected state trajectory and operator models are updated accordingly. This loop repeats until no inconsistencies remain.
After convergence, we obtain a refined fully observable domain $(\mathcal{P}, \mathcal{A})$. When grounded, it provides POMDP components $S$, $A$, $T$, and $R$.

\subsection{Partially Observable Domain Learning}

Next, we derive the partially observable components $(\mathcal{O}, \mathcal{Z})$ by identifying state-observation contradictions.


\subsubsection{Partial Observability Recognition}

We recognize partial observability by identifying mismatches between the reconstructed states and visual states directly parsed from observations.




\begin{enumerate}[
    label=(\alph*),
    wide=0pt,
    leftmargin=0pt,
    labelsep=0.4em,
    itemsep=2pt,
    parsep=0pt,
    topsep=0pt,
    partopsep=0pt
]

\item \textit{Visual state parsing.}
For each step $D_{i,t}$, we parse a symbolic visual state $\hat{s}_{i,t}$, by prompting an LLM to assign predicates in $\mathcal{P}$ according to the scene description $X_{i,t}$,  with values in $\{0,1,?\}$.

\item \textit{Contradiction discovery.}
For each grounded predicate $g=p(\bar{x})$, we compare its value $\hat{s}_{i,t}(g)$ in the visual state with its value $s_{i,t}(g)$ in the reconstructed state. If $\hat{s}_{i,t}(g)\neq ?$ and $\hat{s}_{i,t}(g)\neq s_{i,t}(g)$, we discover a \textit{contradiction} for the predicate schema $p$. 
To avoid mis-detection of contradictions due to erroneous visual grounding or trajectory reconstruction, we use a VLM to check the raw visual observations.
If any mistake is detected, we return to the fully observable learning stage for revision. 

\item \textit{Observable concept formation.}
An LLM proposes an observable schema $o_p$ for each predicate schema $p$ with contradictions discovered; all such schemas form the observable set $\mathcal{O}$. 
Further, we refer to a predicate schema without an associated observable schema as a \textit{fully observable} predicate, since its observation outcomes are always consistent with the hidden state; a predicate schema with an associated observable schema is referred to as a \textit{partially observable} predicate.


\end{enumerate}

\subsubsection{Stochastic Observation Modeling}
Given predicate schemas $\mathcal{P}$ and observable schemas $\mathcal{O}$, we estimate a symbolic observation model $\mathcal{Z}$. 
For fully observable predicates, the observation model is always an identical mapping.
For a partially observable predicate $p$, the observation model specifies $\mathcal{Z}(o\mid p,a)$ and $\mathcal{Z}(o\mid \neg p,a)$, where $o\in\Omega_p=\{o_p^+,o_p^-\}$ denotes a positive or negative observation outcome.

Moreover, observation models are defined differently for manipulation operators and active-perception operators. We refer to them as \textit{passive observation} and \textit{active observation} functions, respectively.

\begin{enumerate}[
    label=(\alph*),
    wide=0pt,
    leftmargin=0pt,
    labelsep=0.4em,
    itemsep=2pt,
    parsep=0pt,
    topsep=0pt,
    partopsep=0pt
]

\item \textit{Passive observation.}
For a step $(i,t)$ corresponding to a manipulation action $a_{i,t}$, we parse the observable values $o_{i,t}$ from the visual state $\hat{s}_{i,t}$, by directly mapping predicate names to the associated observable names. For each partially observable grounded predicate $g=p(\bar{x})$ that exists in $\hat{s}_{i,t}$, we construct a data tuple $\left(o_{i,t}(g),\hat{s}_{i,t}(g),a_{i,t}\right)$.


\item \textit{Active observation.}
The observation model of an active-perception action $a_{i,t}$ highly concentrates on its observation intent, i.e., the predicate for which it aims to gather information.
A VLM watches the raw video clip $(V^h_{i,t},V^w_{i,t})$ to infer which grounded predicate $g=p(\bar{x})$ is of interest, and determines its observation outcome $o_{i,t}(g)\in\Omega_p$. This gives a data tuple $\left(o_{i,t}(g),\hat{s}_{i,t}(g),a_{i,t}\right)$.


\item \textit{Observation likelihood estimation.}
We estimate observation likelihoods by empirical frequency. For each action $a$ and partially observable predicate $p$, samples satisfying $p$ yield the empirical observation distribution $\mathcal{Z}(o\mid p,a)$ for all $o\in\Omega_p$; samples with $\neg p$ yield $\mathcal{Z}(o\mid \neg p,a)$. These conditional distributions are referred to as \textit{observation rules}; the collection of them defines the observation model $\mathcal{Z}$.

\end{enumerate}

The above procedure produces the partially observable components $(\mathcal{O}, \mathcal{Z})$ of the PO-PDDL domain. Once grounded, they induce the POMDP observation space $O$ and observation model $Z$.

\section{Planning with a Learned \formuname Domain}
\label{sec:plan}


Given a learned \formuname domain $\langle \mathcal{P}, \mathcal{O}, \mathcal{A}, \mathcal{Z} \rangle$, we construct a task-specific problem instance. Together with the learned domain, it induces a POMDP over predicate assignments, which we solve via belief-space planning (\figref{fig:pipeline}d).

\subsection{\formuname Problem Generation}
We construct the \formuname problem instance from an initial scene observation and a task instruction $L$, which are used to generate the initial belief $b_0$ and goal condition $\mathcal{G}$, respectively.

\begin{enumerate}[
    label=(\alph*),
    wide=0pt,
    leftmargin=0pt,
    labelsep=0.4em,
    itemsep=2pt,
    parsep=0pt,
    topsep=0pt,
    partopsep=0pt
]

\item \textit{Initial belief generation.}
We construct an initial belief $b_0$ over complete symbolic states. Fully observable predicates are grounded directly from the initial scene observation $(I^{\mathrm{h}}_0, I^{\mathrm{w}}_0)$ using a VLM, yielding deterministic assignments. For partially observable predicates, we identify structural dependencies, such as mutual exclusiveness among object locations, to define a feasible space of consistent assignments. Over this space, we combine an empirical prior $b^{\mathrm{emp}}$ estimated from demonstration initial states $\{s_{i,0}\}$ with a uniform prior $b^{\mathrm{uni}}$ over feasible assignments:
$b_0 = \lambda b^{\mathrm{emp}} + (1-\lambda)b^{\mathrm{uni}}$, where $\lambda\in[0,1]$.

\item \textit{Goal specification.}
The goal set $\mathcal{G}$ is generated from the task instruction $L$ using an LLM. The LLM selects relevant predicate schemas, groups mutually-exclusive predicates, and enumerates feasible assignments within each group. These group-wise assignments are composed into fully-assigned states and checked against $L$; all semantically-matching states are included in $\mathcal{G}$.
\end{enumerate}

\subsection{Belief-Space Planning}

Planning is performed in the belief space over symbolic states induced by the learned \formuname domain. Starting from the initial belief $b_0$, the objective is to select actions that maximize expected cumulative reward under uncertainty, with terminal conditions defined by the goal $\mathcal{G}$.

At each time step $t$, the agent maintains a belief $b_t(s)$ over complete predicate assignments $s$. After executing action $a_t$ and receiving observation $o_{t+1}$, the belief is updated via Bayesian filtering:
$b_{t+1}(s') \propto Z(o_{t+1}\mid s') \sum_{s} T(s'\mid s,a_t)\, b_t(s)$,
where $T$ and $Z$ are induced by the stochastic operator effects and symbolic observation model, respectively.

We solve the resulting POMDP in an online manner using DESPOT~\cite{ye_despot_2017}. At each time step, DESPOT samples state hypotheses from $b_t$, performs sparse belief-tree search by simulating action-observation trajectories under $(T,Z)$, and returns the root action with the highest estimated value. The agent then executes this action, incorporates the new observation, and replans in a receding-horizon loop until sufficient belief mass satisfies $\mathcal{G}$ or termination is reached.

\section{Experiments} 
\label{experiments}



We evaluate our method on real-world robotic planning tasks, to assess whether PO-PDDL can learn reusable symbolic POMDP models under action constraints and stochastic action and perception outcomes. 
Results show that PO-PDDL consistently achieves the strongest planning performance across all tasks, substantially outperforming existing model-learning approaches and VLM planning.


\subsection{Experimental Setup}
\label{experimental_setup}

\begin{wrapfigure}{r}{0.37\linewidth}
    \vspace{-0.8em}
    \centering
    \includegraphics[width=\linewidth]{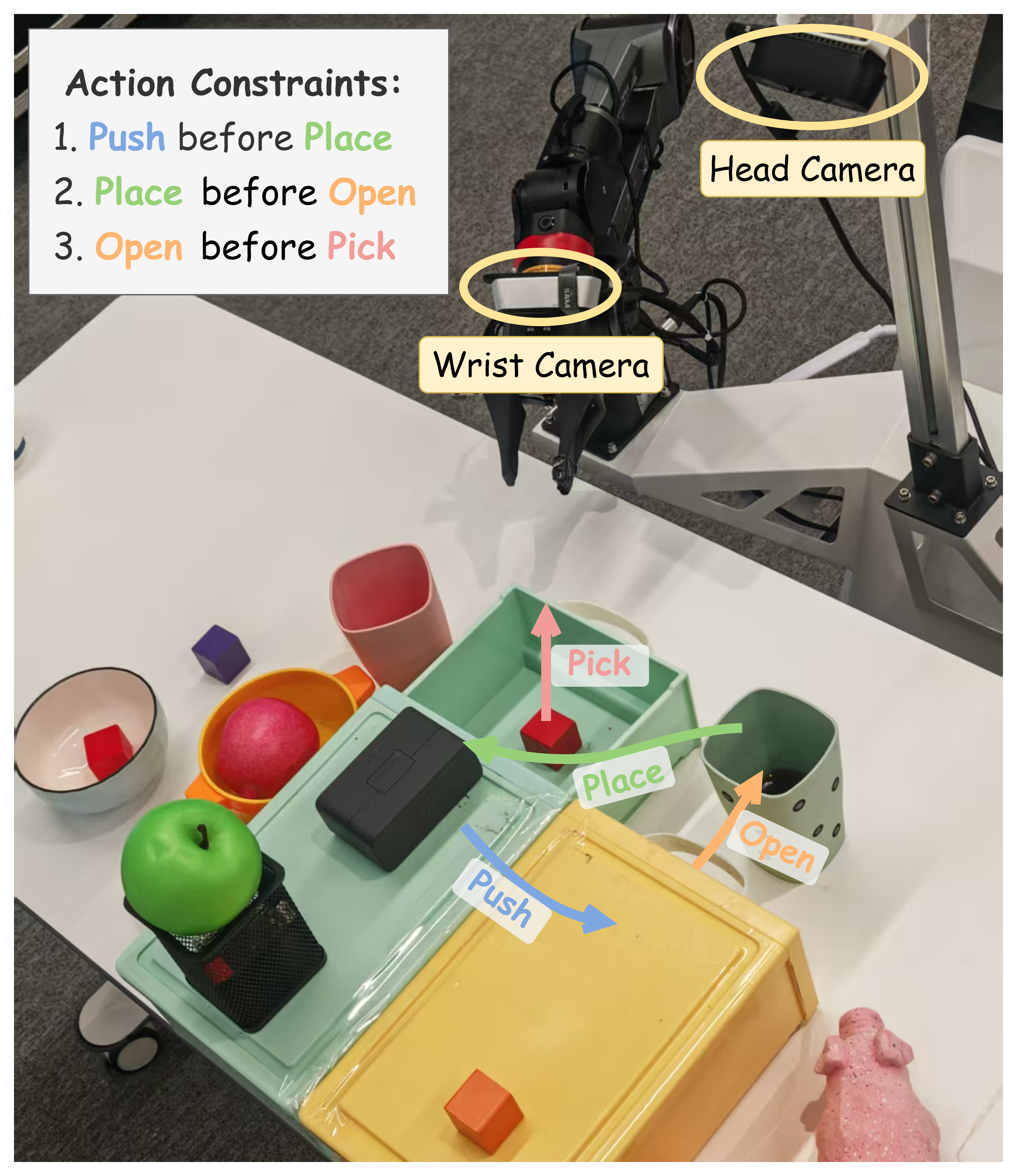}
    \vspace*{-1.5em}
    \caption{Experimental setup.}
    \label{fig:real_world_setup}
    \vspace{-1.5em}
\end{wrapfigure}

We conduct experiments in a real-world tabletop scene, shown in Figure~\ref{fig:real_world_setup}. The environment contains drawers, cups, blocks, and a movable black box. We construct two task sets: (1) 5 short-horizon demonstration tasks (average horizon 3.4 steps), each rolled out multiple times to collect visual demonstrations for domain learning; and (2) 5 longer-horizon evaluation tasks, requiring compositional reasoning and online planning under uncertainty. 

For the low-level stack, manipulation is performed by local VLA skills chained with motion planning, and perception is provided by a VLM operating on RGB observations. 
The VLA policy is based on the public release of $\pi_{0.5}$~\cite{intelligence2025pi_}, post-trained on in-domain robot data. 
We train 41 manipulation skills for opening drawers, moving cups and blocks, and pushing the black box, together with 2 active-perception skills for looking into a cup and looking into a drawer. 
The resulting domain contains both execution uncertainty from imperfect VLA skills and observation uncertainty caused by limited viewpoints, occlusions, hidden interior spaces, and VLM grounding errors.

The five evaluation tasks are designed with increasing difficulty\footnote{For each evaluation task, we assess 10 trials.}:

\begin{itemize}[leftmargin=*, itemsep=0pt, topsep=0pt]
    \item \taskpushbox: ``push the black box to the top of the yellow drawer.'' A 1-step task without partial observability or action constraints, evaluating basic grounding and low-level execution capabilities.

    \item \taskblock: ``put one block on top of one drawer, and close all the drawers.'' The drawer interiors are initially unobservable, while grasping blocks inside drawers exhibits non-negligible failures.

    \item \taskcup: ``place all cups with dark liquid on the top of the green drawer, and keep all cups without dark liquid on the table.'' Cup contents are initially hidden, requiring active perception. A black box occupies the target placement area of cups, requiring constraint resolution.

    \item \taskblockcup: ``put all blocks in the same drawer, and close all drawers.'' A long-horizon task with an average horizon of 7.6 actions under perfect execution. Opening drawers requires first removing cups that block the front, introducing sequential action constraints.

    \item \taskblockcupbox: same task as \taskblockcup, but additionally requiring cups to be placed on top of the green drawer, while the black box initially occupies the target placement area, introducing an additional constraint-resolution step. The average horizon is 8.7 actions under perfect execution.
    
\end{itemize}




We report three evaluation metrics: \textit{task progress}, defined as the fraction of key task elements accomplished; \textit{cumulative reward}, defined by the PO-PDDL reward function; and \textit{planning time}, measured in seconds.
The reward function assigns a $+500$ reward for successful task completion, a $-100$ penalty for falsely declaring task completion, a $-300$ penalty for irreversible failures such as knocking over a cup containing liquid, and a time cost that accounts for both online planning latency and action-execution time. All evaluated methods consistently use GPT-5.4~\cite{OpenAI2026gpt54safety} as the LLM/VLM backend.
See Appendix~\ref{app:experimental_setup_details} for additional details.

\subsection{Baselines}
\label{baselines}

We compare against both explicit model-learning approaches and direct VLM-based planning with history conditioning and in-context learning from demonstrations.

\begin{itemize}[leftmargin=*,itemsep=0pt, topsep=0pt]

    \item \vlmplanner: a VLM directly performs online planning under uncertainty by selecting the next action based on the current visual observation, conditioned on the execution history and language demonstrations (the same demonstrations used for our domain learning).

    \item \unidomain~\cite{yeunidomain}: learns a deterministic PDDL domain from the same visual demonstrations. We adapt its planning module to online planning, by regenerating the PDDL problem at each time step from the current visual observation and solving it with Monte Carlo tree search.

    \item \pomdpcoder~\cite{curtis2025llm}: learns a POMDP model represented as Python code, from language-based demonstrations without visual inputs. We adapt it to our setting by applying our visual grounding pipeline to construct symbolic demonstrations and using our belief initialization and belief update modules for online execution. See Appendix~\ref{app:pomdpcoder_modification} for implementation details.

\end{itemize}

\subsection{Comparison Results}
\label{comparison_results}
\begin{figure}[t!]
    \centering
    \includegraphics[width=0.6\textwidth]{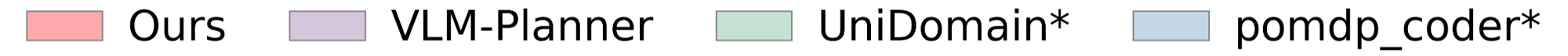}\\[0.5em]
    \begin{tabular}{@{\hskip 2pt}c@{\hskip 2pt}c@{\hskip 2pt}c@{\hskip 2pt}}
        \includegraphics[width=0.33\textwidth]{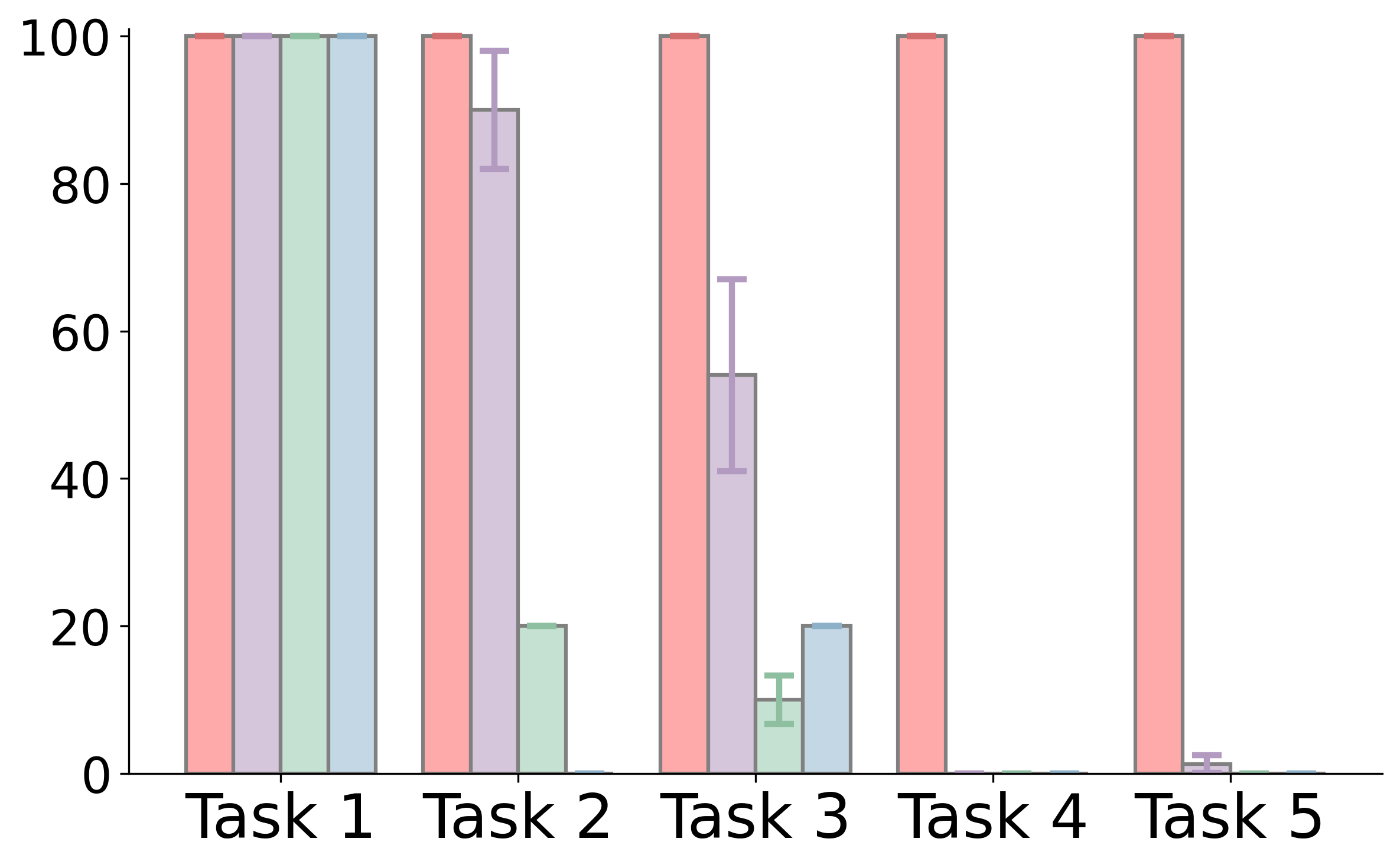} &
        \includegraphics[width=0.33\textwidth]{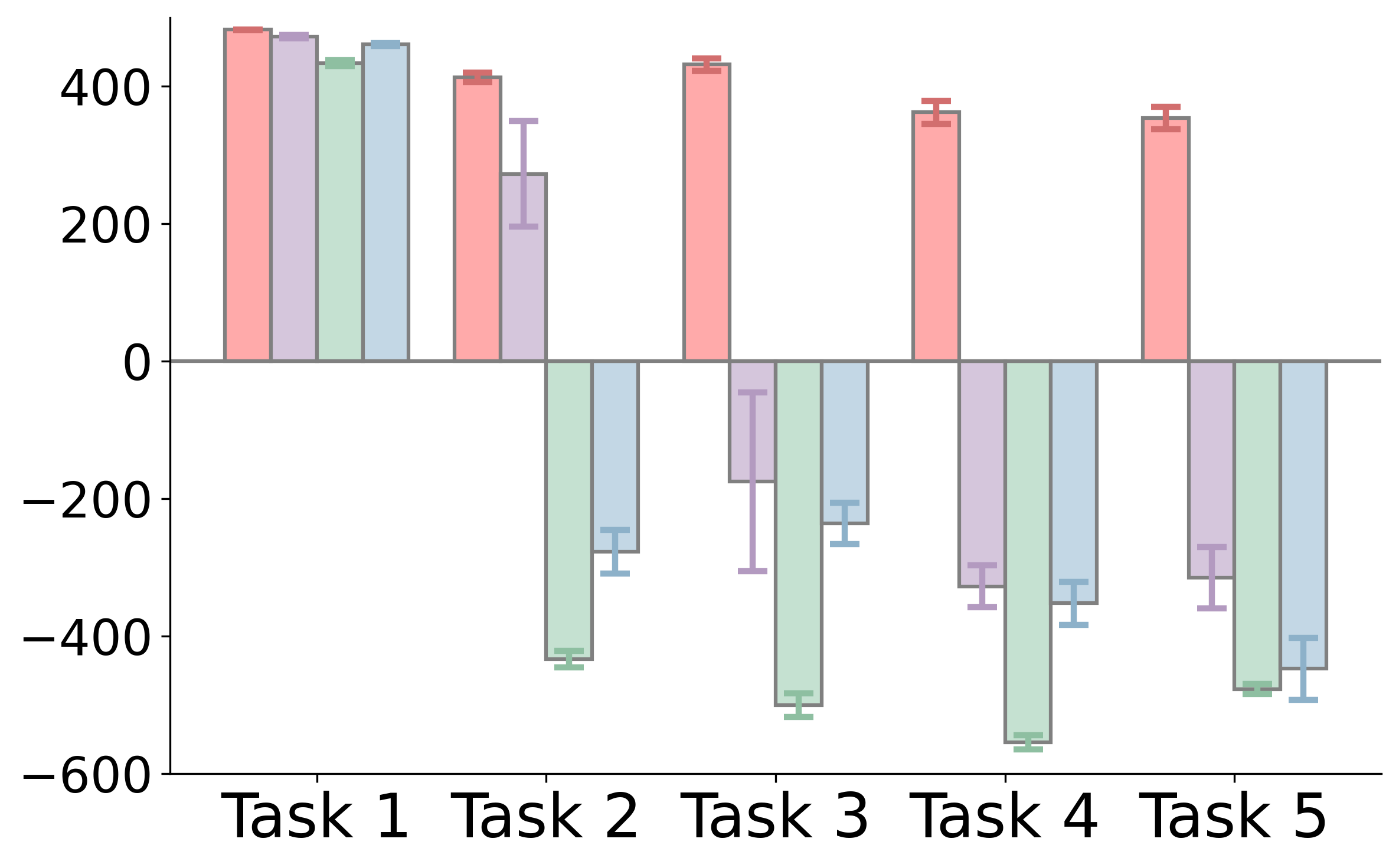} &
        \includegraphics[width=0.33\textwidth]{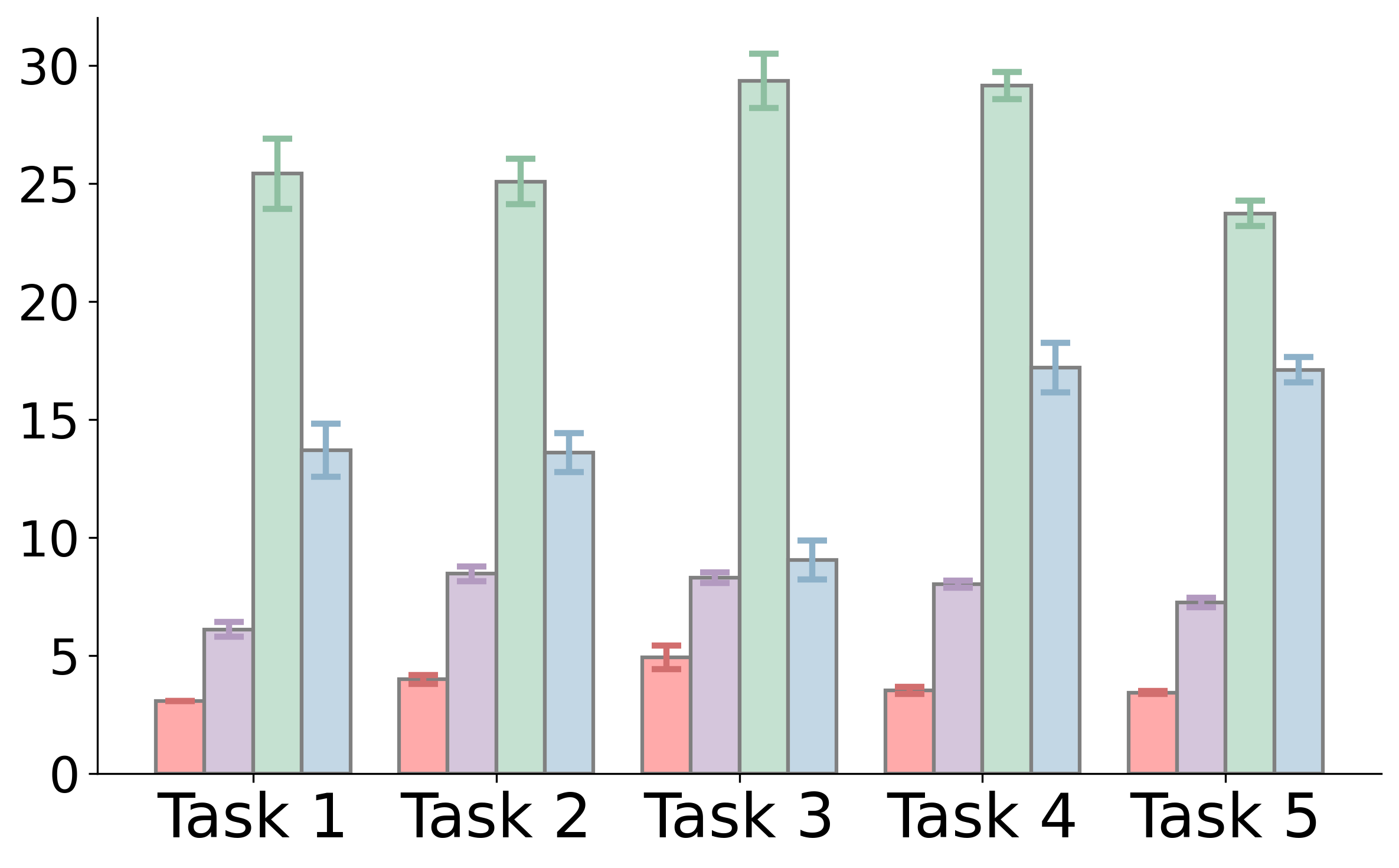} \\
        (a) Task Progress (\%) $\uparrow$ &
        (b) Cumulative Reward $\uparrow$ &
        (c) Planning Time per Step (s) $\downarrow$
    \end{tabular}
    \caption{Performance comparison of our method and baselines. Task 1--5 refer to \taskpushbox, \taskblock, \taskcup, \taskblockcup and \taskblockcupbox respectively.  Each bar represents the average value with standard error (SE).}
    \label{fig:comparison_results}
        \vspace*{-1.0em}

\end{figure}

\paragraph{Domain Learning Results.}

During domain learning, our method learns a correct and complete POMDP model with the lowest learning cost among all explicit model-learning approaches. PO-PDDL takes 291 seconds to learn from 29 demonstrations containing 103 execution steps, while \unidomain and \pomdpcoder require 5111 and 328 seconds, respectively. See Appendix~\ref{app:learning_cost} for detailed learning-time and token-cost statistics. Appendix~\ref{app:learned_model_analysis} further compares the correctness and completeness of the learned models.

\paragraph{Online Planning Results.}

As shown in Figure~\ref{fig:comparison_results}, PO-PDDL consistently outperforms all baselines across all task settings.

\vlmplanner performs reasonably well on \taskpushbox (Task 1) and \taskblock (Task 2), where constraint reasoning is limited. However, it struggles on \taskcup (Task 3), which introduces action constraints and hidden states, and completely fails on the long-horizon tasks \taskblockcup (Task 4) and \taskblockcupbox (Task 5). The VLM often fails to infer the underlying action constraints from visual observations and demonstrations, leading to infeasible actions.

\unidomain performs well on \taskpushbox (Task 1), where uncertainty is minimal and deterministic symbolic planning is sufficient. It achieves only marginal success on \taskblock (Task 2) and \taskcup (Task 3), where partial observability and stochastic execution become important. It completely fails on \taskblockcup (Task 4) and \taskblockcupbox (Task 5), because stochastic execution outcomes and varying pre-states produce inconsistent symbolic representations for the same action, resulting in unsolvable PDDL domains.

\pomdpcoder is generally incapable of modeling complex visual robot domains, and fails on all tasks except the single-step \taskpushbox (Task 1). Although its learned transition and observation programs achieve high coverage of the demonstrated reality, combining them into a globally consistent and solvable POMDP model remains difficult. 

In contrast, PO-PDDL successfully solves all tasks, achieving 100\% task success. The strong performance is attributed to: (1) explicitly modeling action constraints through symbolic operator preconditions; (2) explicitly modeling partial observability and stochastic execution; (3) using an LLM-friendly symbolic formulation for POMDPs; and (4) applying extensive closed-loop verification and revision during domain learning to ensure model correctness and consistency.

Notably, PO-PDDL also achieves the shortest planning time, with an average improvement of 50.4\% over \vlmplanner, 85.7\% over \unidomain, and 73.2\% over \pomdpcoder. This efficiency comes from the fact that most steps only require Bayesian belief updates using the learned transition models. Only actions with observation models require reformulation of the belief with VLMs. 

Prompts of \formuname and all baselines, as well as qualitative results, can be found on the project website: \url{https://roboticsjtu.github.io/PO-PDDL/}.

\subsection{Ablation Studies}
\label{ablation_studies}

We assess the contribution of our key design choices using the following variants:

\begin{itemize}[leftmargin=*,itemsep=0pt, topsep=0pt]

    \item \textit{\jointobslearn}: instead of separating \textit{fully observable domain learning} from \textit{partially observable domain learning}, this variant performs joint learning of both components in a single stage during symbolic grounding and trajectory reconstruction.

    \item \textit{\noactiveobs}: removes the distinction between manipulation operators and active-perception operators during observation modeling. All observations are modeled as passive observations.

\end{itemize}

\begin{wrapfigure}{r}{0.35\textwidth}
    \centering
    \vspace{-0.5em}
    \includegraphics[width=0.35\textwidth]{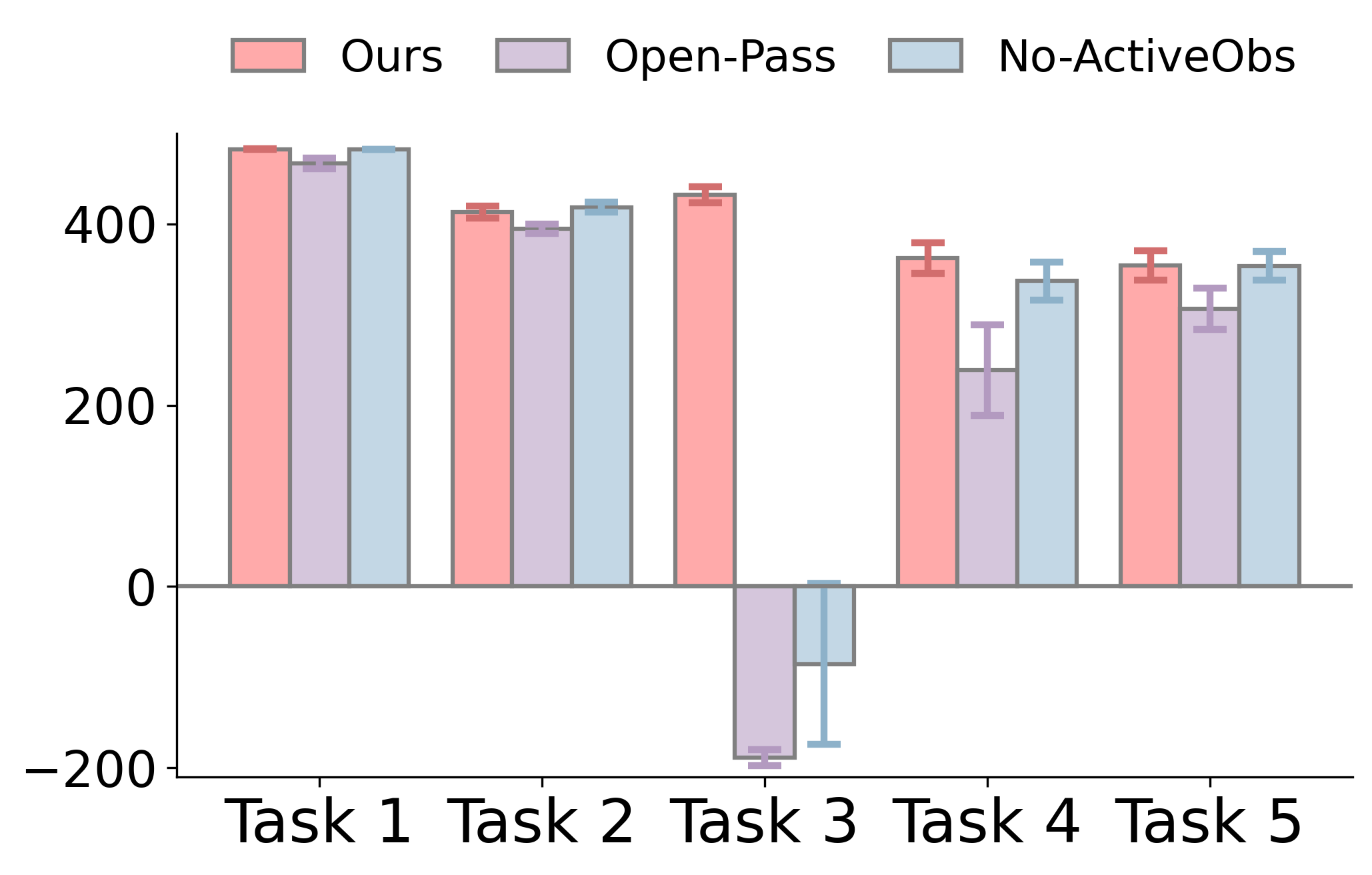}
    \caption{Ablation study results. Values denote cumulative reward.}
    \label{fig:ablation_results}
    \vspace{-1.5em}
\end{wrapfigure}

As shown in Figure~\ref{fig:ablation_results}, both variants exhibit substantially degraded performance.

\textit{\jointobslearn} fails to reliably distinguish fully observable predicates from partially observable predicates, often introducing redundant observables for predicates that provide useless or even destructive information. This produces inefficient belief updates. The issue is particularly severe in \taskcup (Task 3), where unfocused observation modeling corrupts the belief over hidden cup contents, causing planning failures.

\textit{\noactiveobs} under-models active-perception observations. Since passive observations are often consistent with inferred hidden states, the learned model misses the information-gathering role of active-perception operators. In \taskcup (Task 3), this prevents newly observed cup contents from updating the belief, leading to poor task performance.

Beyond these ablations, we further analyze the benefit of online adaptation of operator transition probabilities. See results in Appendix~\ref{online_adaptation}.



\section{Conclusion and Limitations}
\label{conclusion_and_limitations}

We presented \formuname, a symbolic POMDP formulation that extends PDDL with stochastic transitions, partial observability, and belief representations, while preserving its LLM-friendly nature. Building on this formulation, we developed a demonstration-driven pipeline that learns reusable \formuname domains directly from robot execution videos and supports online belief-space planning. Experiments on real-world manipulation tasks show that our method consistently outperforms existing PDDL and POMDP model-learning approaches, achieving substantially stronger planning performance with much lower learning and planning time.

Despite these results, several limitations remain. First, our framework assumes a closed domain without post-demonstration domain shift. Extending \formuname to open-world environments requires online revision of the symbolic POMDP model from new visual observations and interactions. Second, learning cost may grow with large-scale demonstration datasets. A promising direction is to build reusable knowledge bases of modular \formuname domains that can be adapted across scenes and tasks, similar to UniDomain. Finally, system performance remains bounded by the capability and biases of the underlying VLM. Future systems may explicitly identify such bottlenecks and request human assistance when needed.

\newpage
\bibliographystyle{unsrt} 
\bibliography{reference}
\newpage
\appendix
\section{Technical appendices}
\label{appendix}



\subsection{Example \formuname Formulation: Tiger Problem}
\label{app:tiger}

We provide a complete \formuname formulation based on the classic Tiger problem, consisting of a domain file and a problem file. In this problem, an agent faces two doors, with a tiger hidden behind one door and a reward behind the other. The agent can either listen, which provides a noisy observation about the tiger's location, or open one of the doors. Opening the door with the tiger incurs a large penalty, while opening the other door gives a positive reward. After opening a door, the tiger location is reset randomly.

The domain file defines predicates, observable schemas, stochastic operators, rewards, and symbolic observation rules. The problem file specifies the objects, initial belief over hidden symbolic states, goal condition, and optimization metric. Together, the two files instantiate a complete PO-PDDL planning model.

\begin{lstlisting}[style=popddlstyle, caption={PO-PDDL domain file for the Tiger problem.}, label={lst:tiger-domain}]
(define (domain tiger-doors-pomdp)
  (:requirements :strips :typing :probabilistic-effects
                 :conditional-effects :fluents :pomdp)

  (:types
    door
    action-tag
  )

  (:constants
    ;; Spatial convention: two doors arranged left-to-right.
    ;; Names: d_left = left door, d_right = right door.
    d_left d_right - door

    act_listen act_open - action-tag
  )

  (:predicates
    ;; Hidden state: where the tiger is.
    (tiger-at ?d - door)

    ;; Minimal history tag to condition observations on whether we listened last.
    ;; "no-history-yet" is the explicit initial case (no previous action recorded).
    (last-action ?a - action-tag)
    (no-history-yet)
  )

  (:observables
    (obs-hear ?d - door)
    (obs-nothing)
  )

  (:functions
    (total-reward)
  )

  (:action listen
    :parameters ()
    :precondition ()
    :effect (and
      (decrease (total-reward) 1)

      ;; record last action
      (not (no-history-yet))
      (last-action act_listen)
      (not (last-action act_open))
    )
  )

  (:action open
    :parameters (?d - door)
    :precondition ()
    :effect (and
      ;; reward depends on whether tiger is behind the opened door
      (when (tiger-at ?d)
        (decrease (total-reward) 100))
      (when (not (tiger-at ?d))
        (increase (total-reward) 10))

      ;; after opening, reset tiger location uniformly (classic Tiger POMDP reset)
      (not (tiger-at d_left))
      (not (tiger-at d_right))
      (probabilistic
        0.5 (and (tiger-at d_left) (not (tiger-at d_right)))
        0.5 (and (tiger-at d_right) (not (tiger-at d_left)))
      )

      ;; record last action
      (not (no-history-yet))
      (not (last-action act_listen))
      (last-action act_open)
    )
  )

  ;; Observation model: only meaningful after listening; 85% accurate, 15% flipped.
  (:observation hear-when-tiger-left
    :parameters ()
    :condition (and (last-action act_listen) (tiger-at d_left))
    :distribution (probabilistic
      0.85 (obs-hear d_left)
      0.15 (obs-hear d_right)
    )
  )

  (:observation hear-when-tiger-right
    :parameters ()
    :condition (and (last-action act_listen) (tiger-at d_right))
    :distribution (probabilistic
      0.85 (obs-hear d_right)
      0.15 (obs-hear d_left)
    )
  )
)

\end{lstlisting}

\begin{lstlisting}[style=popddlstyle, caption={PO-PDDL problem file for the Tiger problem.}, label={lst:tiger-problem}]
(define (problem tiger-doors-instance)
  (:domain tiger-doors-pomdp)

  (:objects
    ;; Uses domain constants d_left, d_right for the two doors.
  )

  (:init-belief
    ;; Group: tiger-location
    (joint 0.5 (and (tiger-at d_left) (not (tiger-at d_right))))
    (joint 0.5 (and (not (tiger-at d_left)) (tiger-at d_right)))
  )


  (:goal
    ;; Any open action terminates the episode in many formulations; here we model it
    ;; as achieving "last action was open".
    (last-action act_open)
  )

  (:metric maximize (total-reward))
)
\end{lstlisting}

\subsection{Experimental Setup Details}
\label{app:experimental_setup_details}

\textbf{Demonstration Dataset.}
We use 29 short-horizon demonstrations in total, covering five subtask types:
\begin{itemize}[leftmargin=*, itemsep=0pt, topsep=0pt]
    \item Black-box pushing followed by cup placement: ``push the black box to the yellow drawer on the left, and place the \{color\} cup on the top of the green drawer on the right.'' Each demonstration has 3 high-level steps; 4 demonstrations.
    \item Cup placement by liquid content: ``place the cup with/without dark liquid on the top of the green drawer on the right.'' Each demonstration has 3--4 high-level steps; 8 demonstrations.
    \item Cup placement followed by drawer opening: ``place the \{color\} cup on the top of the green drawer on the right, and open the \{color\} drawer.'' Each demonstration has 3 high-level steps; 4 demonstrations.
    \item Block transfer: ``pick the \{color\} block from the \{color\} drawer and place it into/on top of the \{color\} drawer on the left.'' Each demonstration has 3--6 high-level steps; 11 demonstrations. This group includes pick-block failures.
    \item Drawer closing: ``close the \{color\} drawer.'' 2 demonstrations.
\end{itemize}

\textbf{VLA policy and subtask instructions.} We post-trained a $\pi_{0.5}$ model as the last-mile VLA policy for executing manipulation and active-perception skills. All high-level subtask instructions used by the policy are listed below.

\begin{lstlisting}[style=pythonstyle]
TASK_LIST: list[str] = [
    "open the yellow drawer",
    "open the green drawer",
    "close the yellow drawer",
    "close the green drawer",
    "pick up the pink cup on the left of the table",
    "pick up the pink cup on the right of the table",
    "pick up the green cup on the left of the table",
    "pick up the green cup on the right of the table",
    "pick up the pink cup on the top of the green drawer",
    "pick up the green cup on the top of the green drawer",
    "place the pink cup on the top of the green drawer",
    "place the green cup on the top of the green drawer",
    "place the pink cup on the left of the table",
    "place the pink cup on the right of the table",
    "place the green cup on the left of the table",
    "place the green cup on the right of the table",
    "pick up the blue block in the yellow drawer",
    "pick up the blue block in the green drawer",
    "pick up the yellow block in the yellow drawer",
    "pick up the yellow block in the green drawer",
    "pick up the green block in the yellow drawer",
    "pick up the green block in the green drawer",
    "pick up the red block in the yellow drawer",
    "pick up the red block in the green drawer",
    "place the blue block into the yellow drawer",
    "place the blue block into the green drawer",
    "place the yellow block into the yellow drawer",
    "place the yellow block into the green drawer",
    "place the green block into the yellow drawer",
    "place the green block into the green drawer",
    "place the red block into the yellow drawer",
    "place the red block into the green drawer",
    "place the blue block on the top of the yellow drawer",
    "place the blue block on the top of the green drawer",
    "place the yellow block on the top of the yellow drawer",
    "place the yellow block on the top of the green drawer",
    "place the green block on the top of the yellow drawer",
    "place the green block on the top of the green drawer",
    "place the red block on the top of the yellow drawer",
    "place the red block on the top of the green drawer",
    "push the black box on the green drawer to the yellow drawer",
    "look into the cup on the left",
    "look into the cup on the right",
    "look into the yellow drawer on the left",
    "look into the green drawer on the right",
]
\end{lstlisting}

\textbf{Evaluation Tasks and Progress Metrics.}
For each evaluation task, we set a maximum number of execution steps and define a sequence of key progress subtasks for computing task progress. Task progress is measured as the fraction of key progress subtasks achieved during an episode. Unless otherwise specified, these subtasks are checked in the listed order: a later subtask is counted only if all preceding subtasks have been achieved. Subtasks enclosed in braces can be achieved in any order.

\begin{itemize}[leftmargin=*, itemsep=0pt, topsep=0pt]
    \item \taskpushbox. Max steps: 5. Key progress subtasks:
    push the black box to the yellow drawer; report goal completion.
    \item \taskblock. Max steps: 10. Key progress subtasks:
    open one of the drawers; pick one of the blocks; place the same block on top of one of the drawers; close one of the drawers; report goal completion.
    \item \taskcup. Max steps: 10. Key progress subtasks:
    \{look into one of the cups; push the black box\}; pick one of the cups; place the same cup on the green drawer; report goal completion.
    \item \taskblockcup. Max steps: 15. Key progress subtasks:
    pick one of the cups; place the same cup on one of the drawers; open one of the drawers; pick one of the blocks; place the same block in one of the drawers; close one of the drawers; report goal completion.
    \item \taskblockcupbox. Max steps: 15. Key progress subtasks:
    push the black box to the yellow drawer; pick one of the cups; place the same cup on one of the drawers; open one of the drawers; pick one of the blocks; place the same block in one of the drawers; close one of the drawers; report goal completion.
\end{itemize}

\begin{figure}[H]
    \centering
    \begin{subfigure}[t]{0.19\linewidth}
        \centering
        \includegraphics[width=\linewidth]{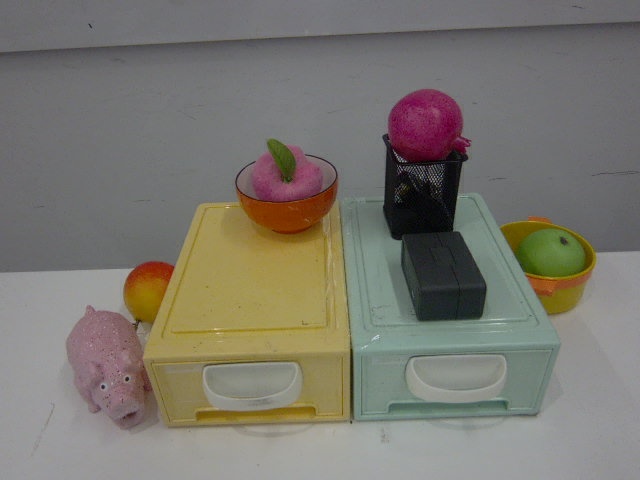}
        \caption{\taskpushbox}
        \label{fig:task1}
    \end{subfigure}
    \hfill
    \begin{subfigure}[t]{0.19\linewidth}
        \centering
        \includegraphics[width=\linewidth]{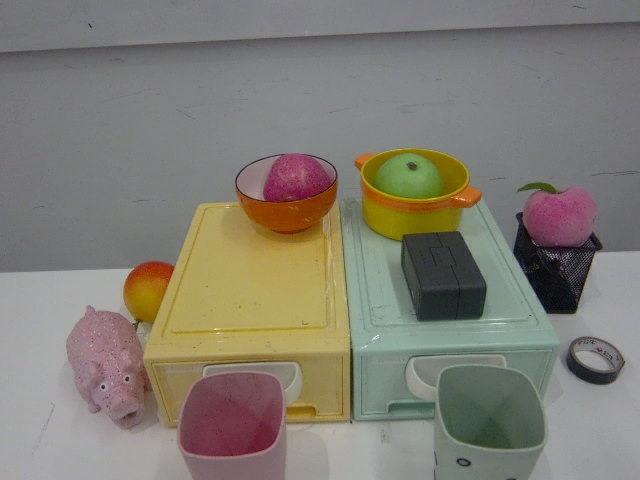}
        \caption{\taskblock}
        \label{fig:task2}
    \end{subfigure}
    \hfill
    \begin{subfigure}[t]{0.19\linewidth}
        \centering
        \includegraphics[width=\linewidth]{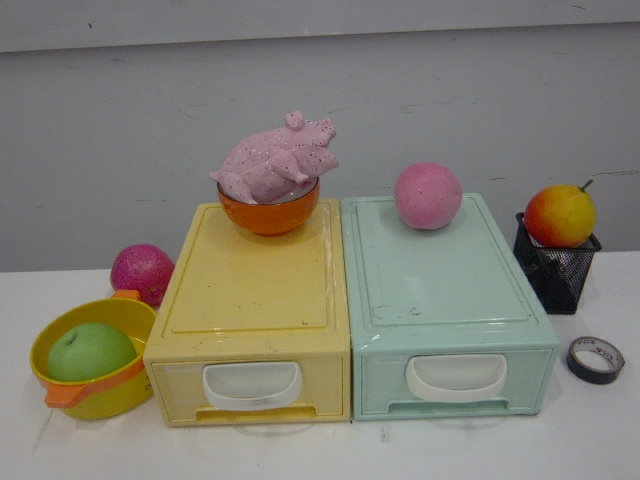}
        \caption{\taskcup}
        \label{fig:task3}
    \end{subfigure}
    \hfill
    \begin{subfigure}[t]{0.19\linewidth}
        \centering
        \includegraphics[width=\linewidth]{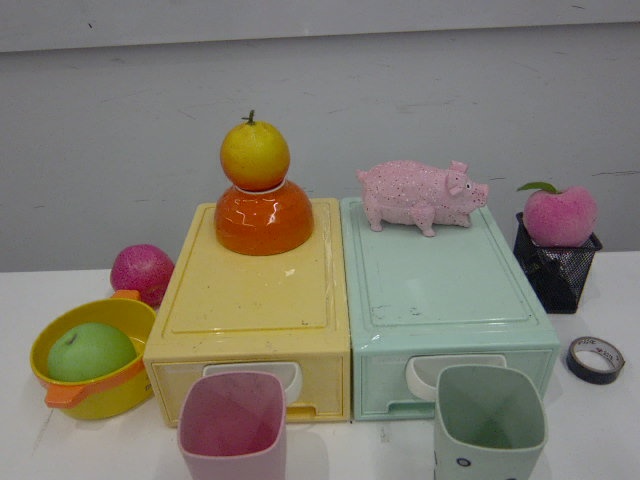}
        \caption{\taskblockcup}
        \label{fig:task4}
    \end{subfigure}
    \hfill
    \begin{subfigure}[t]{0.19\linewidth}
        \centering
        \includegraphics[width=\linewidth]{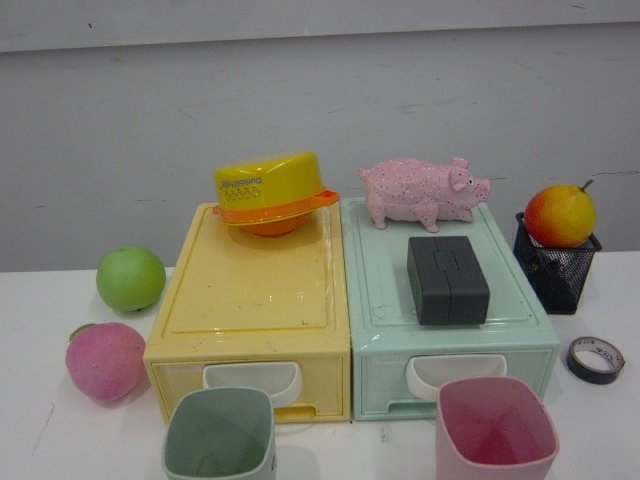}
        \caption{\taskblockcupbox}
        \label{fig:task5}
    \end{subfigure}
    \caption{Real-world experimental setup.}
    \label{fig:tasks}
\end{figure}

\textbf{Initial scene setup.} Figure~\ref{fig:tasks} shows the initial scene configurations for the five evaluation tasks. For each task, we evaluate every method over 10 real-world trials. Across trials, the positions of task-irrelevant distractor objects are randomized. This protocol tests whether each method can robustly solve the same high-level task under small variations in the physical scene, rather than overfitting to a fixed object arrangement.

The task-relevant objects consist of one black box, two blocks (blue and red), two drawers (green and yellow), and two cups (pink and green). For each task: 
\begin{itemize}[leftmargin=*]
    \item \taskpushbox: Both drawers are initially closed, and the black box is placed on top of the green drawer.
    \item \taskblock: Both drawers are initially closed, and the drawer containing each block is randomized across trials.
    \item \taskcup: Both drawers are initially closed and the black box is placed on top of the green drawer, while the positions of the two cups are swapped across trials.
    \item \taskblockcup: Both drawers are initially closed, the drawer containing each block is randomized, and the cup positions are swapped across trials.
    \item \taskblockcupbox: Both drawers are initially closed and the black box is placed on top of the green drawer, while both the block locations and cup positions are varied across trials.
\end{itemize}

\subsection{Modifications for \pomdpcoder Baseline}
\label{app:pomdpcoder_modification}
\subsubsection{Domain-specific Python interface.}
The original \textit{pomdp\_coder} framework learns POMDP components as short probabilistic programs over a domain-specific Python API that defines the state, action, and observation spaces. Therefore, to adapt \textit{pomdp\_coder} to our real-world tabletop manipulation domain, we implement a new Python interface rather than using the APIs from its original benchmark domains. Our interface defines task-native symbolic states, grounded action payloads, symbolic observation payloads, and runtime functions for sampling initial states, transitions, observations, and rewards. This allows \textit{pomdp\_coder} to generate candidate probabilistic programs for our learned manipulation domain while keeping the same high-level model-learning and belief-space planning procedure as the original method.

\begin{lstlisting}[style=pythonstyle]
"""Formal Python interface for learned `pomdp_coder` POMDP models."""

from pathlib import Path
from typing import Any, TypedDict


OBJECT_TYPES_FORMAT = {
    "<type_name>": ["<object_name>", "..."],
}


ACTION_SCHEMA_FORMAT = [
    {
        "name": "<action_name>",
        "parameters": [
            {
                "name": "<parameter_name>",
                "type": "<type_name>",
            }
        ],
    }
]


SPECIAL_ACTION_SCHEMA_FORMAT = [
    {
        "name": "<special_action_name>",
        "parameters": [],
    }
]


STATE_FIELDS_FORMAT = (
    "<state_field_name>",
    "...",
)


OBSERVATION_FIELDS_FORMAT = (
    "<observation_field_name>",
    "...",
)


class ActionPayload(TypedDict):
    """Grounded symbolic action payload."""

    name: str
    args: dict[str, str]


class ObservationPayload(TypedDict):
    """Symbolic observation payload."""

    positive: list[str]
    negative: list[str]
    metadata: dict[str, Any]


class SymbolicState(TypedDict):
    """Task-native symbolic latent state."""

    # Replace these keys with the task-specific fields induced by the task
    # interface. The actual state schema is not fixed globally.
    ...


GOAL_INTERFACE_FORMAT = {
    "type": "<goal_representation_type>",
    "predicate_templates": [
        "<predicate_template>",
        "...",
    ],
}


def sample_initial_state(rng) -> SymbolicState:
    """Sample an initial latent symbolic state."""
    raise NotImplementedError


def sample_next_state(
    state: SymbolicState,
    action: ActionPayload,
    rng,
) -> SymbolicState:
    """Sample the next latent symbolic state."""
    raise NotImplementedError


def sample_observation(
    next_state: SymbolicState,
    action: ActionPayload,
    rng,
) -> ObservationPayload:
    """Sample a symbolic observation emitted after the transition."""
    raise NotImplementedError


def reward(
    state: SymbolicState,
    action: ActionPayload,
    next_state: SymbolicState,
) -> float:
    """Return the scalar reward for the transition."""
    raise NotImplementedError

\end{lstlisting}

\subsubsection{Adaptation Pipeline.}
We adapt \textit{pomdp\_coder} to our manipulation benchmark by preserving its core model-learning principle, while replacing the original benchmark-specific inputs with structured artifacts produced from our visual demonstrations. The original \textit{pomdp\_coder} framework learns the initial-state, transition, observation, and reward components as probabilistic programs, evaluates candidate programs by empirical coverage, and repairs uncovered cases through feedback. In our setting, the key adaptation is therefore to convert our robot demonstrations into the structured symbolic form required by \textit{pomdp\_coder}.

\begin{enumerate}[label=(\alph*), leftmargin=*]
    \item \textbf{Scene-described demonstrations.}
    We start from the same demonstrations used by our method. Each episode is normalized into an annotated trajectory, where every step is associated with a high-level subtask instruction, execution success or failure, and a textual scene description generated from the robot videos. These scene descriptions bridge raw visual inputs and the structured symbolic supervision required by \textit{pomdp\_coder}, mirroring their role in our own pipeline for hidden-state reconstruction and visual-state comparison. Prompting designs are available in the prompt library at \url{https://roboticsjtu.github.io/PO-PDDL/}.

    \item \textbf{Structured data generation.}
    We convert the scene-described demonstrations into a symbolic interface and grounded transition records. The interface defines object types, state fields, action schemas, observation fields, goal representation, reward interface, and special actions, serving as the domain-specific API over which \textit{pomdp\_coder} synthesizes programs. Each annotated episode is then grounded into records containing symbolic pre-state, grounded action, action outcome, symbolic observation, post-state, reward, terminal flag, and metadata. Finally, we construct component-specific specifications for the initial-state, transition, observation, and reward programs, which define the input-output contracts for candidate programs.

    \item \textbf{Candidate program proposal.}
    Given the structured interface and grounded records, we follow \textit{pomdp\_coder}'s proposal stage. The LLM synthesizes multiple candidate Python programs for each POMDP component: the initial-state model, transition model, observation model, and reward model. Each proposal is conditioned on the shared symbolic interface, the component-specific specification, execution rules, a starter template, and a representative subset of grounded training examples. As in our method, the reward component is computed from the empirical average execution time of each action. Using subsets keeps the synthesis problem tractable and encourages compact probabilistic programs rather than memorization of individual demonstrations.

    \item \textbf{Coverage-based evaluation and repair.}
    We retain \textit{pomdp\_coder}'s coverage-based evaluation and repair loop. Candidate programs are executed on grounded symbolic records and evaluated by whether they can reproduce empirical outcomes. For stochastic components, coverage is estimated by repeated sampling: a record is covered if the program can generate the observed symbolic outcome within a fixed Monte Carlo budget. Uncovered examples are converted into mismatch feedback containing the symbolic inputs, expected empirical output, and sampled or predicted output, which is then used by the LLM to repair the full program. The final program is selected by development-set coverage, with mismatch count as a secondary criterion.

    \item \textbf{Runtime integration.}
    After learning the four component programs, we assemble them into a runtime POMDP model and use it in the same online execution setting as our method. The planner maintains a particle-based belief state, propagates particles through the learned transition program, and reweights them using the learned observation likelihood after receiving observations. For a fair comparison, the adapted \textit{pomdp\_coder} baseline uses the same Bayesian belief update and VLM-based observation module as our method. The observation module is invoked only when the learned observation model indicates that the current action may produce informative observations under the current belief. We also retain the Belief-Space Planner used in the original \textit{pomdp\_coder} framework as the POMDP solver.
\end{enumerate}

\subsection{Domain-Learning Cost Comparison}
\label{app:learning_cost}

\textbf{Domain Learning Cost for Our Method and Explicit Baselines.}
Table~\ref{tab:offline_learning_cost} reports the offline domain-learning cost for our method and explicit model-learning baselines on the same 29-demonstration dataset with 103 total high-level steps. Our method achieves the lowest learning time while maintaining a comparable token cost.

\begin{table}[h]
\centering
\caption{Domain-learning Cost Comparison.}
\label{tab:offline_learning_cost}
\resizebox{1.0\linewidth}{!}{
\begin{tabular}{lcccc}
\toprule
Method & Total Time (s)  & LLM/VLM Calls  & Cost-Equiv. Tokens  & Cost-Equiv. Tokens / Step  \\
\midrule
Ours & 291.00 & 541 & 363,247 & 2,752 \\
\pomdpcoder & 328.46 & 201 & 237,042 & 1,796 \\
\unidomain & 5111.19 & 932 & 355,253 & 2,691 \\
\bottomrule
\end{tabular}
}
\vspace{0.2em}

{\footnotesize Cost-equivalent tokens are output-token-equivalent counts computed using the GPT-5.4 API input/output price ratio~\cite{OpenAI2026pricing}: $N_{\mathrm{eq}}=N_{\mathrm{out}}+\frac{1}{6}N_{\mathrm{in}}$. In Tokens / Step, steps include step 0, i.e., the initial observation step.}
\vspace{-0.5em}
\end{table}

Table~\ref{tab:offline_learning_cost} shows that \formuname has the lowest domain learning time among the explicit model-learning methods. \unidomain is substantially slower because its original pipeline extracts atomic domains from demonstrations and then repeatedly fuses them into a unified domain through hierarchical domain merging. This pairwise merging process requires many LLM/VLM calls and is difficult to parallelize, making it costly in our setting. 

\pomdpcoder has a simpler learning structure and fewer total LLM/VLM calls. For fairness, our adaptation also parallelizes many preprocessing and candidate-generation steps, resulting in a slightly higher learning time than ours but lower token usage. However, its weakly constrained Python representation and coverage-based repair do not produce a reliable model for our long-horizon visual manipulation tasks. 

\formuname maintains a reasonable token cost while substantially improving model quality through its structured representation, two-stage domain learning, and explicit observation modeling.

\subsection{Learned Model Analysis}
\label{app:learned_model_analysis}

We analyze the learned models through the six induced POMDP components $\langle S,A,O,T,Z,R\rangle$. The reference is a set of human-verified symbolic requirements derived from the task specification, physical constraints, demonstration annotations, and held-out execution traces. We use this analysis to relate component-level modeling differences to the observed planning behaviors.

\subsubsection{Learned \formuname Model from Our Method}

The learned \formuname model from our method satisfies the required components for the evaluation tasks: it contains task-relevant state and action abstractions, encodes key inter-action constraints (i.e., causal action-ordering constraints), represents stochastic effects, and defines compact action-conditioned observation rules. These components support long-horizon composition under action constraints, stochastic execution, and partial observability, beyond simply covering the short-horizon demonstrations.

\paragraph{State and action spaces.}
The learned state space contains the task-relevant predicates needed to describe object locations, drawer states, gripper states, hidden cup contents, blocking relations, and left/right spatial structure, without introducing redundant overlapping representations. For example, instead of using generic location fields, the learned \formuname model uses typed and side-specific predicates such as \texttt{in\_front\_of\_left}, \texttt{in\_front\_of\_right}, \texttt{on\_left}, \texttt{on\_right}, and side-specific drawer-top predicates. The action space is also side-specific, with separate operators for left-side and right-side manipulation, preserving transition differences caused by side-dependent VLA execution.

\begin{lstlisting}[style=popddlstyle, caption={Structured state and action space in the learned \formuname model.}]
(:predicates
  (contains_dark_liquid ?x0 - cup)
  (gripper_empty)
  (gripper_holding_block ?x0 - block)
  (gripper_holding_cup ?x0 - cup)

  (in ?x0 - drawer ?x1 - block)
  (in_front_of_left ?x0 - cup ?x1 - drawer)
  (in_front_of_right ?x0 - cup ?x1 - drawer)

  (on_left ?x0 - fixed_item)
  (on_right ?x0 - fixed_item)

  (on_top_of_left_block_drawer ?x0 - block ?x1 - drawer)
  (on_top_of_left_box_drawer ?x0 - box ?x1 - drawer)
  (on_top_of_right_box_drawer ?x0 - box ?x1 - drawer)
  (on_top_of_right_cup_drawer ?x0 - cup ?x1 - drawer)

  (open ?x0 - drawer)
)

(:action open_object_on_left ...)
(:action open_object_on_right ...)
(:action pick_up_object_in_object_on_left ...)
(:action pick_up_object_in_object_on_right ...)
(:action place_object_on_top_of_object_on_left ...)
(:action place_object_on_top_of_object_on_right ...)
\end{lstlisting}

This state-action design satisfies the required side-specific spatial distinctions and avoids redundant encodings of the same physical fact. The resulting state space is compact, interpretable, and sufficient for grounding the POMDP state space $S$ and action space $A$.

\paragraph{Preconditions.}
The learned preconditions cover both local executability and verified inter-action constraints. These include constraints such as ``a drawer cannot be opened if a cup blocks its front'' and ``a cup cannot be placed on a drawer top if the black box already occupies that region.'' For example, the learned \texttt{open\_object\_on\_left} operator requires the drawer to be on the left, closed, gripper-empty, and free from blocking cups or objects on relevant drawer-top regions.

\begin{lstlisting}[style=popddlstyle, caption={Precondition encoding drawer-opening constraints.}]
(:action open_object_on_left
  :parameters (?arg0 - drawer)
  :precondition
  (and
    (gripper_empty)
    (not (on_right ?arg0))
    (not (open ?arg0))
    (on_left ?arg0)

    ;; no object held
    (forall (?b - block)
      (not (gripper_holding_block ?b)))
    (forall (?c - cup)
      (not (gripper_holding_cup ?c)))

    ;; no cup blocks the drawer front
    (forall (?c - cup)
      (not (in_front_of_left ?c ?arg0)))
    (forall (?c - cup)
      (not (in_front_of_right ?c ?arg0)))

    ;; no object blocks the relevant drawer-top region
    (forall (?b - block)
      (not (on_top_of_left_block_drawer ?b ?arg0)))
    (forall (?c - cup)
      (not (on_top_of_right_cup_drawer ?c ?arg0)))
  )
)
\end{lstlisting}

This example shows that the learned preconditions encode causal ordering constraints required by long-horizon tasks, rather than only local action applicability. Similarly, the learned \texttt{place\_object\_on\_top\_of\_object\_on\_right} operator excludes boxes on the target drawer top, capturing the need to push the black box away before cup placement.

\paragraph{Effects and stochastic effects.}
The learned effects cover the relevant action outcomes observed in the demonstrations and required by the evaluation tasks. The model supports probabilistic effects by separating different outcomes of the same action into explicit effect modes. For example, picking a block from the left drawer has both a success mode and a failure mode. In the success mode, the robot holds the block and the block is removed from the drawer; in the failure mode, the state does not incorrectly assert that the robot holds the block.

\begin{lstlisting}[style=popddlstyle, caption={Stochastic effect modes for block picking.}]
(:action pick_up_object_in_object_on_left
  :parameters (?arg0 - block ?arg1 - drawer)
  :precondition
  (and
    (gripper_empty)
    (in ?arg1 ?arg0)
    (not (gripper_holding_block ?arg0))
    (not (on_right ?arg1))
    (on_left ?arg1)
    (open ?arg1)
  )
  :effect
  (and
    (decrease (total-reward) 23.536364)
    (probabilistic
      ;; success mode
      0.636364
      (and
        (gripper_holding_block ?arg0)
        (not (gripper_empty))
        (not (in ?arg1 ?arg0)))

      ;; failure mode
      0.363636
      (and
        ;; the block remains in place and the gripper remains empty
      )
    )
  )
)
\end{lstlisting}

The learned effects also maintain state consistency through explicit add/delete updates. For example, the learned box-pushing operator removes the box from the right drawer top and adds it to the left drawer top, rather than updating only one redundant location field.

\begin{lstlisting}[style=popddlstyle, caption={Consistent location update for pushing the box.}]
(:action push_object_on_object_on_right_to_object_on_left
  :parameters (?arg0 - box ?arg1 - drawer ?arg2 - drawer)
  :effect
  (and
    (decrease (total-reward) 21.925)
    (probabilistic
      1.000000
      (and
        (on_top_of_left_box_drawer ?arg0 ?arg2)
        (not (on_top_of_right_box_drawer ?arg0 ?arg1)))
    )
  )
)
\end{lstlisting}

These examples show that the learned transition model $T$ is stochastic and state-consistent. It represents VLA execution uncertainty through effect probabilities and preserves mutually exclusive symbolic locations through explicit add/delete effects.

\paragraph{Observation space.}
The learned observation space is compact and aligned with the verified partially observable predicates. Instead of treating all visible facts as observations, the learned model only introduces observable schemas for predicates that are genuinely uncertain or visually unreliable. In the learned domain, the observation space contains only \texttt{obs\_in} for hidden drawer contents and \texttt{obs\_contains\_dark\_liquid} for cup liquid content, plus \texttt{obs-nothing}. It does not include fully observable facts such as gripper state, drawer open/closed state, or ordinary object locations as redundant observations.

\begin{lstlisting}[style=popddlstyle, caption={Compact learned observation space.}]
(:observables
  (obs-nothing)
  (obs_in ?arg0 - drawer ?arg1 - block)
  (obs_contains_dark_liquid ?arg0 - cup)
)
\end{lstlisting}

This observation space is small, task-relevant, and aligned with the partially observable predicates needed for belief update.

\paragraph{Observation rules.}
The learned observation rules are constructed only for observables and actions that can produce informative or uncertain observations. Passive observation rules are generated for action contexts where visual grounding may be unreliable, while active-perception rules are generated for actions whose purpose is to inspect hidden states. For example, the learned model includes rules for the hidden predicate \texttt{contains\_dark\_liquid}: an initial/passive observation rule captures uncertainty from ordinary visual grounding, while an active-perception rule captures the more informative observation produced by \texttt{look\_into\_cup}.

\begin{lstlisting}[style=popddlstyle, caption={Initial observation rule for cup contents.}]
(:observation initial_contains_dark_liquid_true
  :parameters (?arg0 - cup)
  :condition
  (and
    (contains_dark_liquid ?arg0))
  :distribution 
  (probabilistic
    0.666667 (obs_contains_dark_liquid ?arg0)
    0.333333 (not (obs_contains_dark_liquid ?arg0))
  )
)

(:observation initial_contains_dark_liquid_false
  :parameters (?arg0 - cup)
  :condition
  (and
    (not (contains_dark_liquid ?arg0)))
  :distribution
  (probabilistic
    0.000000 (obs_contains_dark_liquid ?arg0)
    1.000000 (not (obs_contains_dark_liquid ?arg0)))
)
\end{lstlisting}

\begin{lstlisting}[style=popddlstyle, caption={Active-perception observation rule for cup contents.}]
(:observation active_look_into_object_on_left_success_0_contains_dark_liquid_true
  :parameters (?arg0 - cup)
  :condition
  (and
    (last_action_1_param
      look_into_object_on_left_success_0
      ?arg0)
    (contains_dark_liquid ?arg0))
  :distribution
  (probabilistic
    1.000000 (obs_contains_dark_liquid ?arg0)
    0.000000 (not (obs_contains_dark_liquid ?arg0)))
)

(:observation active_look_into_object_on_left_success_0_contains_dark_liquid_false
  :parameters (?arg0 - cup)
  :condition
  (and
    (last_action_1_param
      look_into_object_on_left_success_0
      ?arg0)
    (not (contains_dark_liquid ?arg0)))
  :distribution
  (probabilistic
    0.000000 (obs_contains_dark_liquid ?arg0)
    1.000000 (not (obs_contains_dark_liquid ?arg0)))
)
\end{lstlisting}

This example shows that the learned model does not emit generic visual facts as observations. Instead, it defines action-conditioned observation rules for task-relevant hidden predicates. Ordinary visual grounding provides uncertain evidence about \texttt{contains\_dark\_liquid}, while \texttt{look\_into\_cup} produces a more informative observation for the same hidden state. Thus, the learned observation model $Z$ is tied to specific hidden predicates, observable schemas, and actions that can reveal them.

\subsubsection{Learned PDDL Domain from \unidomain}

 \unidomain recovers many manipulation actions and basic predicates, but several induced POMDP components remain incomplete for the required partially observable and stochastic planning problem. It mainly learns a deterministic PDDL model over local action effects, rather than a model that captures partial observability, action uncertainty, and long-horizon inter-action constraints.

\paragraph{State and action spaces.}
\unidomain learns many basic manipulation predicates, such as object locations, drawer open/closed states, gripper state, and cup liquid content. It also recovers the main manipulation actions for opening and closing drawers, picking and placing objects, and pushing the black box. However, the learned state abstraction does not fully match the required symbolic structure: it misses some task-relevant spatial distinctions and introduces redundant location predicates.

\unidomain lacks explicit left/right localization predicates. Some predicates that should distinguish the left and right sides of the workspace are merged into generic predicates. This also appears in the action schemas, where left-side and right-side operations are often represented by the same action schema. Such merging may be acceptable under deterministic effects, but it loses an important source of transition uncertainty in our setting, because VLA skills can have different success rates on the left and right sides.

\begin{lstlisting}[style=popddlstyle, label={lst:unidomain-merged-location}]
;; UniDomain-style generic location predicates
(on ?o ?x)
(on_drawer ?o ?d)
(in_drawer ?o ?d)

;; Missing side-specific predicates such as
;; (on_left ?d)
;; (on_right ?d)
;; (on_top_of_left_* ...)
;; (on_top_of_right_* ...)
\end{lstlisting}

The learned state space also contains redundant location predicates whose effects are not always kept consistent. For example, \unidomain may represent drawer-top placement using both \texttt{on} and \texttt{on\_drawer}. However, some actions update only one of them. In the following representative effect, pushing an object updates \texttt{on} but not \texttt{on\_drawer}, so the two predicates can become inconsistent.

\begin{lstlisting}[style=popddlstyle, label={lst:unidomain-redundant-state}]
(:action push_to_drawer
  :parameters (?o ?from ?to)
  :precondition (and
    (drawer ?from)
    (drawer ?to)
    (on ?o ?from)
    (hand_free))
  :effect (and
    (not (on ?o ?from))
    (on ?o ?to))
)

;; If (on_drawer ?o ?from) was also true before the action,
;; it is not removed here, producing inconsistent locations.
\end{lstlisting}

\unidomain does not model active perception meaningfully. It learns predicates such as \texttt{inspected} and \texttt{inspected\_inside} for active-perception actions, but these are deterministic action markers. They do not define observable outcomes, observation likelihoods, or belief updates. Therefore, these actions do not provide useful information for planning under partial observability.

\begin{lstlisting}[style=popddlstyle, label={lst:unidomain-inspection-marker}]
(:action look_into_cup
  :parameters (?c)
  :precondition (and (cup ?c))
  :effect (and
    (inspected ?c)
    (inspected_inside ?c))
)

;; This records that inspection happened, but it does not define
;; observations such as obs_contains_dark_liquid or Z(o | s, a).
\end{lstlisting}

\paragraph{Transition model.}
The learned transition model from \unidomain is incomplete for constrained and stochastic manipulation dynamics. It learns local preconditions and deterministic effects for individual actions, but misses three key properties required by the task dynamics: inter-action constraints, location exclusivity, and stochastic failure modes.

\unidomain does not recover important inter-action constraints between actions. For example, placing a cup on top of a drawer should require the target drawer-top region to be unblocked. In the physical domain, the black box may occupy the drawer top, so it must be pushed away before cup placement. However, the learned placement action only checks local conditions such as whether the robot is holding the object, and does not encode this blocking constraint.

\begin{lstlisting}[style=popddlstyle, label={lst:unidomain-place-missing-box}]
(:action place_on_drawer
  :parameters (?o ?d)
  :precondition (and
    (drawer ?d)
    (holding ?o)
    (not (contains_dark_liquid ?o)))
  :effect (and
    (on ?o ?d)
    (on_drawer ?o ?d)
    (hand_free)
    (not (holding ?o)))
)

;; Missing inter-action constraint:
;; the drawer top should be unblocked before placement.
\end{lstlisting}

Similarly, opening a drawer should require the space in front of the drawer to be clear, since a cup in front of the drawer can prevent opening. Picking a block from a drawer should also require the drawer to be open. These constraints are necessary for tasks such as \taskblockcup and \taskblockcupbox, where cups must be moved before drawer opening and drawers must be opened before retrieving blocks. Since \unidomain does not recover these causal ordering constraints, its planner may generate action sequences that are valid in the learned PDDL model but infeasible in the physical scene.

The learned effects do not consistently maintain location exclusivity. Object locations are mutually exclusive in the physical domain: after an object is placed into a new location, its previous location should be removed. In contrast, several \unidomain effects only add the new location predicate without deleting all potentially conflicting old locations. For example, placing an object on a drawer adds \texttt{on} and \texttt{on\_drawer}, but does not remove possible previous location predicates such as \texttt{on\_table}, \texttt{in\_drawer}, or \texttt{at\_drawer\_area}. This can make the same object appear in multiple locations simultaneously in the symbolic state.

\begin{lstlisting}[style=popddlstyle, label={lst:unidomain-location-exclusivity}]
(:action place_on_drawer
  :parameters (?o ?d)
  :precondition (and
    (drawer ?d)
    (holding ?o)
    (not (contains_dark_liquid ?o)))
  :effect (and
    (on ?o ?d)
    (on_drawer ?o ?d)
    (hand_free)
    (not (holding ?o)))
)

;; Missing location-exclusivity effects, e.g.,
;; (not (on_table ?o ?t))
;; (not (in_drawer ?o ?d2))
;; (not (at_drawer_area ?o ?d2))
\end{lstlisting}

This problem is amplified by redundant location predicates in the learned state space. For instance, \unidomain may represent drawer-top placement using both \texttt{on} and \texttt{on\_drawer}. Some effects update only one of these predicates, which can make the two location representations inconsistent. A representative example is pushing an object from one drawer to another: the effect updates \texttt{on}, but does not update \texttt{on\_drawer}.

\begin{lstlisting}[style=popddlstyle, label={lst:unidomain-location-inconsistency}]
(:action push_to_drawer
  :parameters (?o ?from ?to)
  :precondition (and
    (drawer ?from)
    (drawer ?to)
    (on ?o ?from)
    (hand_free))
  :effect (and
    (not (on ?o ?from))
    (on ?o ?to))
)

;; If (on_drawer ?o ?from) was true before pushing,
;; it is not removed, and (on_drawer ?o ?to) is not added.
\end{lstlisting}

\unidomain uses deterministic PDDL effects and therefore cannot represent VLA execution failures. This is problematic because the demonstrations include failed block-picking attempts, and the real robot may fail even when an action is applicable. In a stochastic model, a pick action should have at least two effect modes: a success mode in which the gripper holds the block, and a failure mode in which the block remains in place and the gripper remains empty. The learned deterministic effect can only encode the success branch.

\begin{lstlisting}[style=popddlstyle, label={lst:unidomain-deterministic-effect}]
(:action pick_from_drawer_area
  :parameters (?o ?d)
  :precondition (and
    (drawer ?d)
    (at_drawer_area ?o ?d)
    (hand_free))
  :effect (and
    (holding ?o)
    (not (at_drawer_area ?o ?d))
    (not (hand_free)))
)

;; Missing stochastic effect modes:
;; success: holding the object
;; failure: object remains in place and gripper remains empty
\end{lstlisting}

Thus, \unidomain learns a deterministic local transition model rather than the stochastic, constraint-aware, and state-consistent transition model required for the tasks. It can describe some successful action effects in isolation, but it does not preserve mutually exclusive object locations, encode long-horizon action-ordering constraints, or model execution failures.

\paragraph{Reward and planning objective.}
Because \unidomain is based on standard PDDL, it does not learn a reward function or action-cost model comparable to the $R$ component in the induced POMDP. In our online adaptation, we use Monte Carlo tree search with a uniform action prior at each step. The search is performed over deterministic symbolic effects from the learned PDDL model. As a result, \unidomain can solve simple fully observable tasks, but it fails to produce robust and cost-effective behavior in long-horizon tasks with action uncertainty and partial observability.

\subsubsection{Learned POMDP Model from \pomdpcoder}

Compared with \unidomain, \pomdpcoder explicitly learns Python programs for the initial-state model, transition model, observation model, and reward model. However, the learned model is weakly structured: state variables, action semantics, transition effects, and observations are implemented through dictionaries and procedural rules rather than explicit relational predicates, stochastic operator modes, and action-conditioned observation rules.

\paragraph{State and action spaces.}
The learned state space is loose and redundant relative to the required symbolic state structure. Instead of complete assignments over grounded predicates, \pomdpcoder represents states using a small set of dictionaries, such as \texttt{object\_location}, \texttt{drawer\_state}, \texttt{object\_properties}, and \texttt{support\_occupancy}. This representation can encode many relevant variables, but it does not enforce relational structure, typing, or mutual-exclusion constraints as explicit symbolic requirements. It also lacks explicit left/right localization predicates. For example, side information is mostly encoded through object names such as \texttt{green\_drawer} and \texttt{yellow\_drawer}, rather than predicates such as \texttt{on\_left}, \texttt{on\_right}, \texttt{in\_front\_of\_left}, and \texttt{in\_front\_of\_right}.

\begin{lstlisting}[style=pythonstyle, caption={Learned state representation.}]
state = {
    "holding": None,
    "object_location": {},
    "drawer_state": {},
    "object_properties": {},
    "support_occupancy": {
        "table_top": "unknown",
        "green_drawer_top": None,
        "yellow_drawer_top": None,
    },
    "goal_spec": [],
}
\end{lstlisting}

This representation also introduces redundant state variables. For instance, an object on a drawer top can be represented both by \texttt{object\_location[obj] = "on\_top\_of(green\_drawer)"} and by \texttt{support\_occupancy["green\_drawer\_top"] = obj}. Such redundancy requires every transition effect to update multiple fields consistently. A structured symbolic model should maintain such consistency explicitly through predicate-level effects or equivalent constraints. The action space has a similar limitation: as in \unidomain, left-side and right-side manipulation are not explicitly separated, which loses transition differences caused by side-dependent VLA execution success rates.

\paragraph{Transition model.}
The learned transition model partially improves over \unidomain in local precondition modeling. For example, it checks whether an object is reachable before picking it, and requires the target drawer to be open before placing an object inside. However, these preconditions are implemented as scattered Python \texttt{if} statements. Since the program structure is loose, the LLM must enumerate all necessary spatial, temporal, and causal preconditions procedurally. This is difficult to do completely from short-horizon demonstrations, leading to missing constraints in long-horizon planning.

\begin{lstlisting}[style=pythonstyle, caption={Learned transition preconditions.}]
if name == "place_in":
    obj = args.get("obj")
    dst = args.get("dst")
    if obj is None or dst is None:
        return next_state
    if holding != obj:
        return next_state
    if drawer_state.get(dst) != "open":
        return next_state
    _place_in_container(next_state, obj, dst)
    return next_state

if name == "open_container":
    dst = args.get("dst")
    if dst is None:
        return next_state
    drawer_state[dst] = "open"
    _maybe_reveal_object_on_open(next_state, dst)
    return next_state
\end{lstlisting}

The first branch shows that \pomdpcoder can learn useful local preconditions, such as requiring an open drawer for \texttt{place\_in}. However, the \texttt{open\_container} branch directly opens the drawer without checking whether a cup blocks the drawer front. This misses an important inter-action constraint required by \taskblockcup and \taskblockcupbox: cups must be moved away before the corresponding drawer can be opened.

The learned effects show mixed quality. Compared with \unidomain, \pomdpcoder attempts to maintain location consistency through helper functions such as \texttt{\_clear\_previous\_support}. However, this consistency is maintained procedurally and remains incomplete. More importantly, the model lacks a clear effect-mode structure that would guide the learner to separate different outcomes of the same action, such as success and failure. As a result, action failures are not modeled as explicit stochastic effect modes, even though VLA execution is stochastic.

\begin{lstlisting}[style=pythonstyle, caption={Learned transition effects.}]
def _clear_previous_support(next_state, obj):
    support_occupancy = next_state.setdefault("support_occupancy", {})
    object_location = next_state.setdefault("object_location", {})
    old_loc = object_location.get(obj)
    if isinstance(old_loc, str) and old_loc.startswith("on_top_of("):
        container = old_loc[len("on_top_of("):-1]
        top_key = _container_top_key(container)
        if support_occupancy.get(top_key) == obj:
            support_occupancy[top_key] = None

def _place_on_top(next_state, obj, container):
    _clear_previous_support(next_state, obj)
    next_state["holding"] = None
    next_state.setdefault("object_location", {})[obj] = \
        f"on_top_of({container})"
    next_state.setdefault("support_occupancy", {})[
        _container_top_key(container)] = obj
\end{lstlisting}

This helper-based implementation is more careful than the deterministic PDDL effects learned by \unidomain, but it is still ad hoc. The consistency of \texttt{object\_location} and \texttt{support\_occupancy} depends on whether every action calls the right helper and updates all related fields. In addition, some effects mix state transition with information revelation. For example, opening a drawer calls \texttt{\_maybe\_reveal\_object\_on\_open}, which assigns every object with unknown location to the opened drawer. This turns an observation event into a physical state transition, corrupting the latent state rather than updating the belief.

\begin{lstlisting}[style=pythonstyle, caption={Information reveal encoded as transition.}]
def _maybe_reveal_object_on_open(next_state, container):
    object_location = next_state.setdefault("object_location", {})
    for obj, loc in list(object_location.items()):
        if loc == "unknown":
            object_location[obj] = f"in_container({container})"
\end{lstlisting}

Although \pomdpcoder can express stochastic programs, the learned transition model remains weakly structured. Its preconditions are incomplete because they must be enumerated as Python conditions, and its effects lack an explicit stochastic effect-mode structure for modeling action failures.

\paragraph{Observation space and observation model.}
\pomdpcoder has a more complete observation interface than \unidomain: it explicitly returns symbolic observation payloads with positive observations, negative observations, and metadata. However, its observation space is overly broad and redundant. It mixes fully observable facts with partially observable facts, including gripper holding state, drawer open/closed state, visible object locations, object properties, and support-clear observations. The required observation space should instead focus on task-relevant partially observable predicates, with each observable tied to a specific hidden predicate.

\begin{lstlisting}[style=pythonstyle, caption={Learned observation payload.}]
return {
    "positive": sorted(positive),
    "negative": sorted(negative),
    "metadata": metadata,
}
\end{lstlisting}

The learned observation function further illustrates this redundancy. It emits observations for many ordinary visible facts, such as holding state and drawer state, even though these are not necessarily the task-relevant hidden predicates that require belief updates.

\begin{lstlisting}[style=pythonstyle, caption={Learned observation model.}]
# Holding is almost always visible.
if holding is None:
    positive.add("holding(null)")
    positive.add("holding_null")
    for obj in objects:
        negative.add("holding(%s)" % obj)
else:
    positive.add("holding(%s)" % holding)
    for obj in objects:
        if obj != holding:
            negative.add("holding(%s)" % obj)

# Drawer states are generally observable when known.
for container in containers:
    ds = drawer_state.get(container, "unknown")
    if _drawer_observable(ds, action_name):
        if ds == "open":
            positive.add("drawer_open(%s)" % container)
            negative.add("drawer_closed(%s)" % container)
        elif ds == "closed":
            positive.add("drawer_closed(%s)" % container)
            negative.add("drawer_open(%s)" % container)
\end{lstlisting}

This observation space lacks an explicit binding between an observable outcome and the hidden predicate it is meant to inform. As a result, the observation model behaves more like a procedural visibility heuristic than a compact action-conditioned likelihood model. It also does not clearly model which observable instances an active-perception action is intended to query. Instead, active perception is mixed with generic visibility rules over object locations, drawer states, and object properties.

\begin{lstlisting}[style=pythonstyle, caption={Generic visibility-based observations.}]
# Visible object locations.
for obj in objects:
    if _visible_object(next_state, obj, action_name):
        visible_objects.append(obj)
        loc = object_location.get(obj, "unknown")
        for atom in _location_atoms(obj, loc):
            positive.add(atom)
        _add_negatives_for_object(negative, obj, loc, containers)

# Support occupancy sometimes yields clear(top) observations.
for support_name, occ in support_occupancy.items():
    if isinstance(support_name, str) and support_name.endswith("_top"):
        container = support_name[:-4]
        if occ is None:
            if action_name in ("push_to", "place_on_top", "pick") \
                    or rng.random() < 0.4:
                positive.add("clear(%s)" % support_name)
                for obj in objects:
                    negative.add("on_top_of(%s, %s)" % (obj, container))
\end{lstlisting}

These issues arise from the same source as the transition-model errors: the model is learned as a loose Python program and does not use a two-stage learning pipeline that reconstructs a reliable fully observable domain before identifying truly partially observable predicates. Consequently, \pomdpcoder tends to produce redundant observations and may invoke VLM-based observation more frequently during online execution.

\paragraph{Reward and planning implications.}
For the reward component, \pomdpcoder uses the same empirical average action duration as our method, rather than autonomously inducing a distinct reward model. Thus, the main differences are not in $R$, but in the learned $S$, $A$, $T$, $O$, and $Z$. Overall, \pomdpcoder learns a valid POMDP-style Python interface and can cover short demonstration records, but its loose state representation, incomplete procedural preconditions, weak effect-mode structure, and redundant observation model make it unreliable for long-horizon belief-space planning. This explains why it can solve the simplest task but generalizes poorly to longer-horizon tasks with action constraints and partial observability.

\subsection{Qualitative Experiment: Online Adaptation of Operator Effect Probabilities}
\label{online_adaptation}
Our structured \formuname representation allows operator effect probabilities to be updated during online execution. Since each outcome can be mapped to a symbolic effect mode, the planner can recalibrate transition probabilities from real feedback without relearning the domain structure.

\begin{figure}[H]
    \centering
    \includegraphics[width=1.0\linewidth]{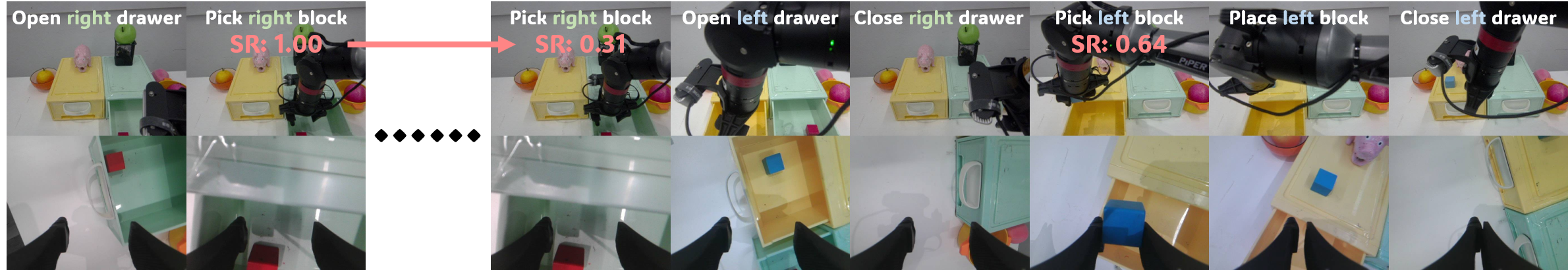}
    \caption{An example of online adaptation on \taskblock. SR: success rate for the action.}
    \label{fig:online_adaptation}
\end{figure}

Figure~\ref{fig:online_adaptation} shows a qualitative example on \taskblock. To simulate test-time skill degradation from domain shift, we make right-drawer grasping unreliable during evaluation. The learned model initially prefers the right drawer, since it learned a lower success probability for left-drawer grasps from demonstrations. After repeated failures, the estimated success probability of right-drawer grasping decreases, and the planner switches to opening the left drawer and grasping the block from there. 

The adapted probabilities also provide a diagnostic signal for low-level skill weaknesses. For example, when transferring to a new domain, a skill whose estimated success probability drops repeatedly, such as right-drawer grasping in Figure~\ref{fig:online_adaptation}, indicates a mismatch between the learned model and the current execution environment. These low-probability or rapidly degrading skills can be prioritized for targeted data collection, additional demonstrations, or VLA fine-tuning, instead of uniformly collecting more data for all skills.


\subsection{Hyperparameters in Online POMDP Planning}
\label{hyper_parameters}

\begin{table}[H]
\centering
\begin{tblr}{
    width = \textwidth,
    colspec = {Q[1.5cm,c,m] X[c,m] Q[1cm,c,m]},
    hlines, vlines,
    row{1} = {font=\bfseries},
    cells = {font=\small}
}
\textbf{Parameter} & \textbf{Description} & \textbf{Value} \\ 
$\lambda$ & Empirical--uniform prior mixing weight for initial belief generation.  & 0.5 \\
$T$ & Temperature for all LLMs and VLMs & 0.0 \\
$k$ & Number of scenarios in Belief Tree Search & 500 \\
$d_s$ & Maximum search depth in Belief Tree Search & 50 \\
$d_r$ & Rollout policy execution depth & 20 \\
\end{tblr}
\end{table}



\end{document}